\pdfoutput=1
\newif\ifsubmit
\submittrue

\documentclass[12pt]{article}
\usepackage[letterpaper, margin=1.05in]{geometry}
\providecommand{\keywords}[1] {\noindent\textbf{{Keywords:}} #1.}
\usepackage[backend=biber, style=numeric, sorting=nyt, maxbibnames=10, maxalphanames=10, minalphanames=10]{biblatex} 
\let\localcite\cite
\renewcommand{\cite}[1]{\localcite{#1}}
\let\localciteauthor\citeauthor
\newcommand{\citet}[1]{\localciteauthor{#1}~\localcite{#1}}

\usepackage[T1]{fontenc}
\usepackage[english]{babel}
\usepackage{csquotes}

\usepackage{mathtools}
\usepackage{amssymb}

\usepackage[T1]{fontenc}
\usepackage[english]{babel}
\usepackage{csquotes}
\usepackage{xspace}

\usepackage{mathtools}

\usepackage{amsthm}
\newtheorem{definition}{Definition}[section]

\newtheorem*{example*}{Example}
\theoremstyle{remark}

\usepackage{graphicx} \graphicspath{{./figures/}}
\usepackage{standalone}
\usepackage{tikz}
\usetikzlibrary{arrows.meta,fit,positioning}
\usepackage{cuted}
\usepackage{pgfplots}
\usepackage{pgfplotstable}
\pgfplotsset{compat=1.18}
\usepgfplotslibrary{groupplots}

\usepackage{enumitem}
\usepackage{booktabs}
\usepackage{array, multirow}
\usepackage{diagbox}
\usepackage{makecell}
\setcellgapes{2.5pt}

\usepackage{url} \PassOptionsToPackage{hyphens}{url} 

\usepackage{hyperref} \PassOptionsToPackage{breaklinks}{hyperref}
\usepackage{xcolor}
\hypersetup{colorlinks, linkcolor={red!50!black}, citecolor={blue!50!black}, urlcolor={blue!80!black}}
\usepackage[ruled, linesnumbered]{algorithm2e}

\usepackage[capitalise]{cleveref}
\crefname{definition}{Definition}{Definitions}
\crefname{proposition}{Proposition}{Propositions}
\crefname{theorem}{Theorem}{Theorems}
\crefname{corollary}{Corollary}{Corollaries}
\crefname{example}{Example}{Examples}
\crefname{remark}{Remark}{Remarks}

\usepackage[symbols, acronym, nonumberlist, nogroupskip, stylemods={mcols,longbooktabs}, section=subsection, numberedsection]{glossaries-extra} \setabbreviationstyle[acronym]{long-short} \glssetcategoryattribute{symbol}{nohyper}{true} \glssetcategoryattribute{acronym}{nohyper}{true}
\makenoidxglossaries

\newcommand{\paragraphNoSkip}[1]{\paragraph{#1.}}

\newcommand{\R}{\mathbb{R}}

\newacronym{ap}{AP}{average precision}
\newacronym{bn}{BN}{batch normalization}
\newacronym{ci}{CI}{confidence interval}
\newacronym{cv}{CV}{cross-validation}
\newacronym{ecdf}{ECDF}{empirical cumulative distribution function}
\newacronym{knn}{kNN}{$k$-nearest neighbors}
\newacronym{lmbfgs}{LM-BFGS}{limited-memory Broyden-Fletcher-Goldfarb-Shanno}
\newacronym{mlp}{MLP}{multilayer perceptron}
\newacronym{ntxent}{NT-Xent}{normalized temperature-scaled cross-entropy}
\newacronym{pca}{PCA}{principal component analysis}
\newacronym{rbf}{RBF}{radial basis function}
\newacronym{relu}{ReLU}{rectified linear unit}
\newacronym{rocauc}{ROC-AUC}{area under the receiver operating characteristic curve}
\newacronym{svd}{SVD}{singular value decomposition}
\newacronym{svm}{SVM}{support vector machine}
\newacronym{tfidf}{TF-IDF}{term frequency-inverse document frequency}

\glsxtrnewsymbol[description={Set of observed accounts}]{accounts}{\ensuremath{\mathcal U}}
\glsxtrnewsymbol[description={Account $u$'s ordered event history}]{history}{\ensuremath{E_u}}
\glsxtrnewsymbol[description={Number of events in $E_u$}]{eventcount}{\ensuremath{n_u}}
\glsxtrnewsymbol[description={Number of temporal views formed from $E_u$}]{viewcount}{\ensuremath{m_u}}
\glsxtrnewsymbol[description={The $j$-th temporal view of account $u$}]{view}{\ensuremath{E_u^{(j)}}}
\glsxtrnewsymbol[description={Feature map from a history to $\mathbb R^{d_{\mathrm{in}}}$}]{featuremap}{\ensuremath{x(\cdot)}}
\glsxtrnewsymbol[description={Shared account encoder with parameters $\theta$}]{encoder}{\ensuremath{f_\theta}}
\glsxtrnewsymbol[description={Full-history representation $f_\theta(x(E_u))$}]{representation}{\ensuremath{h_u}}
\glsxtrnewsymbol[description={Training projection head with parameters $\psi$}]{projector}{\ensuremath{g_\psi}}
\glsxtrnewsymbol[description={Task classifier with parameters $\phi$}]{classifier}{\ensuremath{q_\phi}}
\glsxtrnewsymbol[description={Binary task label of account $u$}]{label}{\ensuremath{y_u}}
\glsxtrnewsymbol[description={Contrastive temperature}]{temperature}{\ensuremath{\tau}}

\newcommand{\system}{{\textsc{FISSION}}\xspace}

\title{{\system: Label Augmentation for Bot Detection}}
\author{Sen Yang\textsuperscript{1} \and Ignacy Nieweglowski\textsuperscript{2} \and Aviv Yaish\textsuperscript{1}}
\date{\small
\textsuperscript{1}Yale University, IC3\\
\textsuperscript{2}Staples High School}
\begin{document}
\maketitle
\begin{abstract}
Bot accounts and coordinated influence operations are often discovered via heuristic methods, leaving a dearth of reliable ground-truth labels for training detection systems.
To address this challenge, we study a natural question: can we generate labels to assist in learning embeddings in which bots and accounts from the same coordinated operation are close?
We present FISSION, a method to generate labels by splitting each account's activity into positively labeled sub-accounts.
Given this label source, we train detection models which preserve behavioral regularities recurring across positive sub-accounts.
We evaluate FISSION and show it outperforms prior methods in detecting Wikipedia sockpuppets and Twitter/X bots.

\end{abstract}

\section{Introduction}
Platforms try to recognize abusive accounts despite scarce and aging labels.
Obtaining reliable labels can require expert investigation, whereas labels derived from account-collection heuristics may encode artifacts of the collection process itself.
Indeed, high within-dataset performance can fail to transfer when, e.g., the platform or the time period changes \cite{hays2023simplistic}.
This problem is especially difficult in Sybil and coordinated-abuse settings, where accounts can be intentionally constructed to obscure shared control and behavioral similarity.

Yet an account's history contains a source of supervision that requires no external annotation: two disjoint portions of the same history have a common observed origin.
For example, interaction patterns and lexical choices may vary across individual events while retaining regularities across these portions.
This leads to a natural question:
\begin{quote}
\emph{Can an account's own history provide the supervision needed to learn representations for abuse detection when labels are scarce?}
\end{quote}

\paragraphNoSkip{This Work}
We answer this question with \system, which aligns views of the same account history without seeing an abuse label.
A classifier fitted on a small labeled sample then maps the resulting account representation to the task class.

We evaluate \system on the Wikipedia sockpuppet dataset of \citet{raszewski2025detecting} and the Twitter/X dataset of \citet{cresci2015fame}.
For Wikipedia, \system reaches $98.684\%$ accuracy, $81.957\%$ sockpuppet $F_1$, and $94.543\%$ \gls{rocauc}.
On Cresci15, an \gls{rbf} \gls{svm} \cite{cortes1995support} reaches $99.227\%$.
The improvement is greatest when labels are scarce, as we show by varying the amount of labels.
Thus, $1\%$ of the Wikipedia training accounts puts \system more than $6\%$ above raw features, while $5\%$ of Cresci15 labels brings it within $1\%$ of the full-label results.

We proceed like so.
First, we formulate within-history origin as a source of ground-truth augmentation and instantiate it through ``fissioned'' views.
Then, we provide account-disjoint evaluations on two platforms, including matched raw-feature controls and label-efficiency experiments.
Next, we perform a component-level analysis of \system's gains: On Wikipedia, most of the improvement is already present with one encoder.
On Cresci15, the measured improvement appears at the level of the full representation stack, while the contribution of count-adaptive view construction varies with the downstream classifier.

\begin{figure*}[t]
\centering
\includegraphics[width=0.95\textwidth]{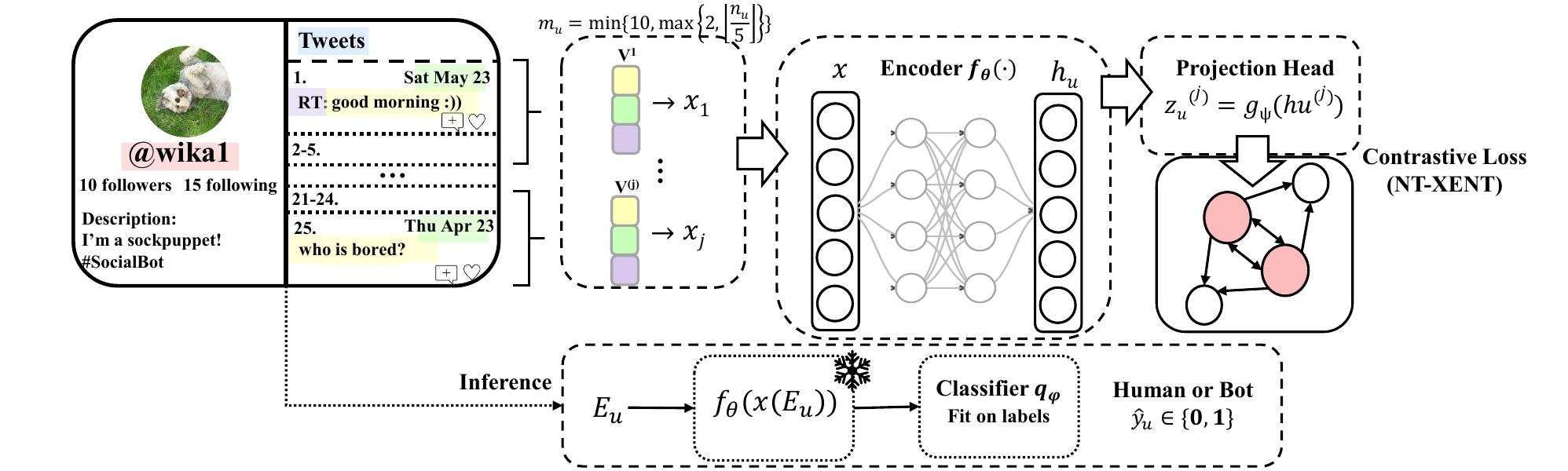}
\caption{\system samples positive pairs from disjoint temporal views of the same unlabeled history.
The \glsentryshort{ntxent} objective trains encoder $f_\theta$ through a temporary projection head $g_\psi$.
Classification discards $g_\psi$, retains $f_\theta$, and uses labels only to fit $q_\phi$.}
\label{figPipelineSample}
\end{figure*}

\begin{table*}[t]
\centering
\caption{Comparison with prior work on the sockpuppet dataset.
[RD25] refers to \cite{raszewski2025detecting}.
}
\label{tabCompareSockpuppet}
\resizebox{\textwidth}{!}{
\begin{tabular}{lccccccc}
\toprule
\textbf{Approach} & \textbf{AUROC} & \textbf{AUPRC / AP} & \textbf{F1-Score} & \textbf{F0.5-Score} & \textbf{Accuracy} & \textbf{Precision} & \textbf{Recall}
\\
\midrule
Random [RD25] & $50.10 \pm 0.14$ & $50.85 \pm 0.09$ & $40.34 \pm 0.11$ & $36.52 \pm 0.12$ & $50.10 \pm 0.16$ & $34.46 \pm 0.12$ & $50.05 \pm 0.15$
\\
Majority [RD25] & - & - & - & - & $65.60 \pm 0.00$ & - & -
\\
RoBERTa [RD25] & $65.70 \pm 0.00$ & $50.45 \pm 0.03$ & $57.97 \pm 0.01$ & $57.52 \pm 0.06$ & $66.98 \pm 0.06$ & $59.54 \pm 0.13$ & $67.63 \pm 0.17$
\\
Standard Enc. [RD25] & $68.33 \pm 0.09$ & $50.67 \pm 0.33$ & $60.05 \pm 0.18$ & $58.73 \pm 0.16$ & $68.90 \pm 0.07$ & $59.72 \pm 0.32$ & $69.88 \pm 0.34$
\\
Pre-trained Enc. [RD25] & $62.74 \pm 0.02$ & $44.80 \pm 0.19$ & $57.49 \pm 0.13$ & $52.90 \pm 0.12$ & $62.79 \pm 0.05$ & $51.45 \pm 0.25$ & $74.76 \pm 0.28$
\\
Reptile Enc. [RD25] & $78.98 \pm 0.12$ & $62.21 \pm 0.08$ & $67.46 \pm 0.53$ & $67.89 \pm 0.17$ & $77.51 \pm 0.19$ & $69.43 \pm 0.26$ & $70.81 \pm 0.82$
\\
\midrule
\textbf{Our Work} & \textbf{94.543($\pm$0.005)} & \textbf{84.084$\pm$0.012} & \textbf{81.957($\pm$0.025)} & \textbf{87.182$\pm$0.057} & \textbf{98.684($\pm$0.002)} & \textbf{91.052$\pm$0.084} & \textbf{74.515$\pm$0.027}
\\
\bottomrule
\end{tabular}
}
\end{table*}

\begin{table}[t]
\centering
\caption{Comparison with prior work on Cresci15.
}
\label{tabCompareCresci15}
\begin{tabular}{ccc}
\toprule
\textbf{Method} & \textbf{Accuracy} & \textbf{F1-score}
\\
\midrule
\citet{hu2020heterogeneous} & 97.45($\pm$0.23) & 96.87($\pm$0.43)
\\
\citet{lv2021are} & 96.84($\pm$0.13) & 97.44($\pm$0.74)
\\
\citet{feng2022heterogeneity} & 97.15($\pm$0.32) & 97.78($\pm$0.24)
\\
\citet{feng2022botrgcn} & 96.52($\pm$0.71) & 97.30($\pm$0.53)
\\
\citet{lei2023bic} & 98.35($\pm$0.24) & 98.71($\pm$0.18)
\\
\citet{liu2023botmoe} & 98.50($\pm$0.00) & 98.82($\pm$0.00)
\\
\citet{cai2024lmbot} & 99.06($\pm$0.30) & 99.26($\pm$0.20)
\\
\citet{li2025ets} & 99.10($\pm$0.08) & 99.29($\pm$0.06)
\\
\citet{zhang2026rabot} & 99.14($\pm$0.21) & 98.94($\pm$0.34)
\\
\midrule
\textbf{Our Work} & \textbf{99.227($\pm$0.061)} & \textbf{99.389($\pm$0.048)}
\\
\bottomrule
\end{tabular}
\end{table}

\section{Related Work}

\paragraph{Account-Abuse Detection.}
Social-bot detection has moved from hand-crafted account signals toward learned representations.
SATAR learns an account embedding directly, while BotRGCN and BotMoE incorporate graph structure and multiple modalities \cite{feng2021satar,feng2021botrgcn,liu2023botmoe}.
Wikipedia sockpuppet detection has followed a parallel path from stylometry toward behavioral and contextual models whose labels commonly originate in platform investigations \cite{solorio2013case,sakib2022automated,raszewski2025detecting}.
These systems demonstrate the value of account representations, although supervised accuracy can also reward collection shortcuts that fail to generalize across datasets \cite{hays2023simplistic}.
\system instead learns its encoder from within-history structure and uses investigation labels only for the final classifier.

\paragraph{Self-Supervised User Representations.}
Contrastive learning defines a representation through the information its selected views are asked to share \cite{chen2020simclr,saunshi2019theoretical}.
Temporal approaches contrast contexts or subsequences to preserve regularities that recur at more than one time scale \cite{yue2022ts2vec}.
User-representation methods exploit the same continuity across observations of one identity \cite{gu2021behavioral,wupasum2024,guo2026somer}.
\system retains the canonical objective but forms positives by partitioning the entire history into disjoint, activity-adaptive views.
Each event is used once, short histories remain dense, and prediction returns to the complete history.
We test whether this relation improves account classification under matched label budgets.

\paragraph{Coordination, Linkage, and Sockpuppets.}
Coordination methods compare accounts for evidence that their actions share a controller \cite{pacheco2021coordinated,sharma2021identifying}.
\system begins from a relation available within each observed history: two disjoint periods have the same source.
Thus, we use that relation to learn account representations.

\section{Problem Setting and \system}

\subsection{Account Histories and Task Labels}

\begin{definition}[Account-Classification Problem]
\label{defAccountClassificationProblem}
Let $\mathcal U_{\mathrm{tr}}$ and $\mathcal U_{\mathrm{te}}$ be disjoint sets of training and test accounts.
Account $u$ has an ordered history $E_u=(e_{u1},\ldots,e_{un_u})$, where each event has a timestamp and observed platform attributes.
A labeled subset $\mathcal L\subseteq\mathcal U_{\mathrm{tr}}$ provides $y_u\in\{0,1\}$.
The goal is to learn a score for $\mathcal U_{\mathrm{te}}$, whose labels are used only for evaluation.
\end{definition}

Representation learning receives the histories of $\mathcal U_{\mathrm{tr}}$ but not task labels $y_u$ or links between accounts.
The feature map $x$ sends a history or temporal view to $\R^{d_{\mathrm{in}}}$, and we seek an encoder $f_\theta:\R^{d_{\mathrm{in}}}\to\mathbb S^{d-1}$ for which a classifier $q_\phi(f_\theta(x(E_u)))$ can be estimated from a small labeled set.
This separates information sources: every eligible training history may shape $f_\theta$ through its structure, while only $\mathcal L$ determines the task boundary learned by $q_\phi$.

\subsection{Temporal Fission}

\begin{definition}[Temporal Fission]
\label{defTemporalFission}
An $m$-fission of $E_u$ is an ordered tuple $\mathcal F_m(E_u)=\bigl(E_u^{(1)},\ldots,E_u^{(m)}\bigr)$ of $m \in 2, \dots, n_u$ nonempty contiguous subsequences of lengths differing by at most one whose concatenation recovers $E_u$.
\end{definition}

Temporal fission preserves contiguity and assigns every event to exactly one period of the history.
The feature map summarizes each period, and the shared source of any two views is known from their construction.
Unlike overlapping or independently sampled windows, a fission covers the history once, so a positive pair cannot agree merely because the same event appears in both views.

\begin{definition}[Common-History Relation]
\label{defCommonHistoryRelation}
For two views $E_u^{(i)}$ and $E_v^{(j)}$, their common-history label is $r_{\mathrm{hist}}\!\left(E_u^{(i)},E_v^{(j)}\right)=\mathbf 1[u=v]$.
The positive-pair set for account $u$ is:
$$
\mathcal P_u
=
\left\{\left(E_u^{(i)},E_u^{(j)}\right): 1 \le i<j\le m_u\right\}
.
$$
\end{definition}

Because the relation is constructed within each history, every eligible training account supplies positive pairs without a task label.
For histories eligible for contrastive training, we choose the number of views from activity volume: \begin{equation}
m_u=\min\!\left\{10,\max\!\left\{2,\left\lfloor n_u/5\right\rfloor\right\}\right\}.
\label{eqViewCount}
\end{equation}
The resulting $\binom{m_u}{2}$ unordered pairs supply contrastive examples.
\Cref{eqViewCount} avoids dividing short histories into ten nearly empty pieces while retaining several temporal comparisons for active accounts.
Before the ten-view cap is reached, each adaptive view contains about five events.
Beyond the cap, additional activity makes the views denser without increasing the 45-pair maximum.
Thus, the rule changes temporal resolution where histories are short and limits pair growth where histories are long.
At inference, the encoder receives $x(E_u)$ for the complete history, including histories too short to contribute training pairs.

\subsection{Behavioral and Lexical Views}

Each temporal view is mapped to a compact account-level feature vector.
For Cresci15, 39 dimensions combine activity histograms with interaction, text-length, and inter-post summaries.
Wikipedia uses 30 dimensions for when an account edits and how edits distribute across pages, sizes, comments, and time.
\Cref{secFeaturesAndFissionedViews} gives the feature set.

Within each outer fold, we reduce 5,000 \gls{tfidf} features \cite{salton1988term} to 50 dimensions by truncated \gls{svd} \cite{deerwester1990indexing}.
Smoothed inverse document frequency downweighs common terms and $\ell_2$ normalization controls for document length.
Truncated \gls{svd} gives a low-rank latent semantic basis.
The transforms remain fixed before evaluation.
With the behavioral dimensions, they give 89 inputs on Cresci15 and 80 on Wikipedia.
Normalization emphasizes behavioral composition over activity volume and preserves one schema for views and histories.

\subsection{Contrastive Representation Learning}

The encoder is a \gls{mlp} \cite{rumelhart1986learning} with widths $d_{\mathrm{in}}\!\rightarrow128\rightarrow128\rightarrow64$.
Its hidden layers use \gls{bn} and \gls{relu}, and its output is normalized to the unit sphere \cite{ioffe2015batch,nair2010rectified}.
Following the literature (e.g., SimCLR \cite{chen2020simclr}), a $64\!\rightarrow128\!\rightarrow32$ projection network applies the contrastive \gls{ntxent} loss and is discarded after pretraining.

In a $2B$ view minibatch, $v_i$ denotes view $i$, $z_i=g_\psi(f_\theta(x(v_i)))$ is its normalized projected representation, and $p(i)$ indexes its paired view.
We minimize the \gls{ntxent} loss:
\begin{equation}
\mathcal L_{\mathrm{NTX}}
=-\frac{1}{2B}\sum_{i=1}^{2B}
\log
\frac{\exp(z_i^\top z_{p(i)}/\tau)}
{\sum_{k\ne i}\exp(z_i^\top z_k/\tau)}.
\label{eqNtxent}
\end{equation}
We train for 50 epochs with Adam at learning rate $10^{-3}$, batches of $2^8$ pairs and temperature $\tau=0.12$ \cite{kingma2015adam}.
Unit normalization makes the dot products cosine similarities, the paired view supplies the numerator, and every other view in the minibatch enters the denominator.
Temperature controls how strongly the objective concentrates on the most similar in-batch alternatives.
Each materialized pair appears in both directions, so either view serves once as anchor and once as its designated positive.
The loss is a single-positive approximation to the common-history relation: if another pair from the same account enters the minibatch, its views remain in the denominator.
\Cref{secFeaturesAndFissionedViews} reports the corresponding masking check.
Materializing $\mathcal P_u$ gives account $u$ a weight proportional to $\binom{m_u}{2}$, so adaptive views couple activity volume to pair count.
The fixed ten-way comparison gives every eligible account 45 pairs.
We combine independently fitted encoders by averaging their class probabilities, a standard ensemble construction \cite{dietterich2000ensemble}.

\begin{definition}[Account Prediction Ensemble]
\label{defAccountPredictionEnsemble}
For $K$ pretrained encoders, account $u$ has representations $\forall k \in \left[K\right]: h_u^{(k)}=f_{\theta_k}(x(E_u))$.
If $q_{\phi_k}(h)\in[0,1]$ is the estimated positive-class probability from the classifier paired with encoder $k$, the ensemble prediction is:
$$
\widehat p_u
=
\frac{1}{K}\sum_{k=1}^{K}q_{\phi_k}\!\left(h_u^{(k)}\right)
.
$$
\end{definition}

We use $K=3$ and fit one classifier to each 64-dimensional representation.
Probability averaging reduces dependence on any one fitted encoder without increasing the representation dimension seen by a classifier.
\Cref{secClassifierFamilyRobustness} compares this with a classifier on the concatenated representation.
The fixed-view model is a learned-representation variant, while subsequent comparisons add adaptive segmentation and probability averaging.

\subsection{Why the Relation Can Help}

Why should a relation about identity help a classifier predict abuse?
Aligning a positive pair rewards behavior that recurs across the account's history.
The representation helps when those recurring traits also support the downstream task.
More views provide more positive pairs but make each summary noisier.
\Cref{eqViewCount} keeps sparse histories in denser views while capping the contribution of highly active accounts.
Pretraining can then use every eligible training history, while labels are needed only to fit the downstream classifier.

\paragraph{Computational Profile.}
The cap $m_u\le10$ limits each account to 45 materialized pairs.
Training therefore scales with $\sum_u\binom{m_u}{2}$, while the minibatch supplies cross-account negatives without enumerating the quadratic set of account pairs.
At inference, each feature vector passes through the encoder once and produces a 64-dimensional representation.
The three-encoder model repeats this operation and averages the classifier probabilities.

\section{Experimental Design}

\subsection{Questions and Datasets}

The central empirical question is whether common-history pretraining improves matched account classification as task labels disappear.
Controlled component and feature comparisons then locate any difference.

\begin{table*}[t]
\centering
\caption{The cohorts that on which our system is evaluated experimentally.
Wikipedia's active cohort supplies the \system teacher and the controlled representation analyses.
Sparse histories in the full cohort are scored by \system-Distill.}
\label{tabCohorts}
\resizebox{\textwidth}{!}{
\begin{tabular}{@{}llrrrc@{}}
\toprule
\textbf{Dataset} & \textbf{Positive Class} & \textbf{Positive} & \textbf{Negative} & \textbf{Total} & \textbf{View Eligibility}
\\
\midrule
Cresci15 & Bot & 3,351 & 1,950 & 5,301 & At least 20 events
\\
Wikipedia Active & Sockpuppet & 66,247 & 95,357 & 161,604 & At least 10 edits
\\
Wikipedia Full & Sockpuppet & 132,890 & 3,179,197 & 3,312,087 & Routed by edit count
\\
\bottomrule
\end{tabular}}
\end{table*}

\paragraph{Wikipedia.}
We use an English Wikipedia dataset whose labels derive from public sockpuppet investigations \cite{solorio2013case,sakib2022automated,raszewski2025detecting}.
After blank identifiers are removed and conflicting row labels are resolved by a positive-wins rule, the full scoring cohort contains 3,312,087 account identifiers (\cref{tabCohorts}).
Retaining accounts with at least ten edits yields the 161,604-account active cohort used for encoder training and controlled representation analyses.
The dataset records account, time, page, recorded edit size, comment, and label, but it does not identify investigations or master accounts.
The activity filter raises sockpuppet prevalence from $4.0\%$ in the full scoring cohort to $41.0\%$ in the active cohort.

\paragraph{Full-Dataset \system-Distill.}
To extend the active-cohort signal to sparse histories, frozen out-of-fold \system probabilities supervise label-free students on chronological prefixes.
A robust-scaled logistic student handles one-edit accounts, edit-count-specific XGBoost regressors handle accounts with two through nine edits, and accounts with at least ten edits retain the frozen \system teacher score \cite{chen2016xgboost}.
The routed scores receive account-disjoint, cross-fitted Platt calibration and positive-$F_1$ threshold selection: each account is excluded from its own calibrator and threshold fit, although labels from the other folds are used at these two downstream stages.
Student scores are sealed before sparse-account task labels are read.
\cref{secWikipediaFullCoverage} gives the complete protocol and route-level diagnostics.

\paragraph{Cresci15.}
We use the public Twitter dataset of \citet{cresci2015fame}.
All 5,301 profile accounts are evaluated.\footnote{Released post rows cover 5,148 accounts.
The remaining 153 are routed to the profile-only fallback described in \cref{secNoPostFallback}.
``Without released posts'' does not imply zero lifetime posts.}
For accounts covered by released events, only training accounts with at least 20 events form fission pairs.
Shorter histories remain part of the evaluation and are fully embedded.

\begin{figure*}[t]
\centering
\includegraphics[width=0.95\textwidth]{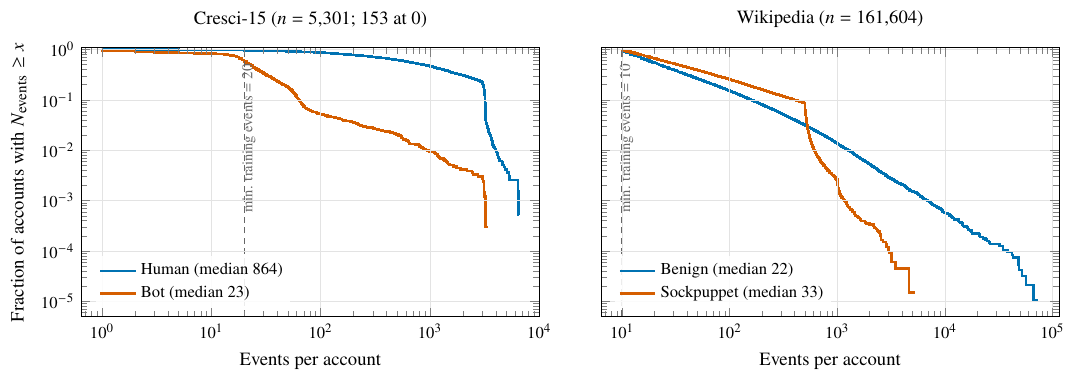}
\caption{Activity support in two evaluated cohorts.
The median Cresci15 human has 864 released events, compared with 23 for a bot.
Wikipedia is less polarized, with medians of 22 edits for benign accounts and 33 for sockpuppets.
Curves are empirical complementary cumulative distributions, with each account counted once.
Dashed lines mark the minimum history length for contrastive training.
The Cresci15 curves cover all 5,301 profiles: the 153 profiles without released post rows are counted at zero in the empirical distribution and fall outside the logarithmic horizontal axis.
Every retained Wikipedia account is eligible.}
\label{figPostDistributions}
\end{figure*}

\subsection{Account-Disjoint Model Selection}

\begin{definition}[Account-Disjoint Outer Evaluation]
\label{defAccountDisjointOuterEvaluation}
For every outer fold $f$, let $\mathcal U_{\mathrm{tr}}^{(f)}\cap\mathcal U_{\mathrm{te}}^{(f)}=\varnothing$.
The evaluation is account-disjoint when no event of any $u\in\mathcal U_{\mathrm{te}}^{(f)}$ is used to fit the feature transform, representation, classifier, calibration map, or threshold that scores $u$.
\end{definition}

\begin{table}[t]
\centering
\caption{Information available at each outer evaluation stage.}
\label{tabLabelAccess}
\makegapedcells
\begin{tabular}{@{}>{\raggedright\arraybackslash}p{0.23\columnwidth}
>{\raggedright\arraybackslash}p{0.7\columnwidth}@{}}
\toprule
\textbf{Stage} & \textbf{Information Used for Fitting}
\\
\midrule
Feature Transforms & Complete histories in outer fold training portion, no task labels
\\
Encoder Pretraining & Eligible fissioned training views, no task labels
\\
Classifier Selection & Representations of labeled training accounts and inner-fold validation labels
\\
Outer Evaluation & Complete held-out histories and labels used only for scoring
\\
\bottomrule
\end{tabular}
\end{table}

We repeat stratified five-fold \gls{cv} 10 times on Cresci15 and 3 times on the Wikipedia active cohort.
Within every outer fold, a three-fold inner \gls{cv} selects classifier hyperparameters and the threshold that maximizes positive-class $F_1$ \cite{cawley2010overfitting}.
This nested design separates model selection from evaluation, avoiding the optimistic bias that arises when both use the same folds.
We provide reproducibility details in \Cref{secCompleteExperimentalProtocol}.

The Wikipedia active-cohort experiment uses $\ell_2$-regularized logistic regression, with $\Pr_\phi(y=1\mid h)=\sigma(w^\top h+b)$ and $\sigma(a)=(1+\exp(-a))^{-1}$.
Class weighting is selected in the inner folds.
Cresci15 uses \gls{rbf} \gls{svm}, whose kernel $k(h,h')=\exp(-\gamma\lVert h-h'\rVert_2^2)$ is monotone in cosine similarity for unit-normalized representations \cite{cortes1995support}.
Its decision values are converted to probabilities by a training-only Platt map \cite{platt1999probabilistic}.
We evaluate logistic regression on Cresci15 to hold the classifier family fixed.
So, the linear and kernelized downstream classifiers evaluate if learned representations are effective across classifier families.

Raw-feature controls share the accounts, training-fold transforms, labeled subsets, inner-fold selection, and threshold rule used by \system.
The controls standardize the raw vectors within each training partition and use the same fitted lexical basis as the learned-representation models.
Cresci15 raw controls use both logistic and \gls{rbf} \gls{svm} classifiers.

\subsection{Controls and Component Tests}

\Cref{tabExperimentalContrasts} summarizes what changes and what is held fixed in each comparison.
The central contrast fixes the accounts, features, logistic classifier, folds, and threshold rule while replacing the raw vector with one fixed-view representation.
Other contrasts examine adaptive views, probability averaging across encoders, classifier family, feature removal, and label budget.
Fixed ten-way fissioning gives every eligible account 45 pairs, separating the central comparison from the activity-dependent pair counts induced by \cref{eqViewCount}.

Feature ablations retrain the complete model after removing an input group.
View construction is compared separately within each classifier family, while comparisons across classifier families reuse the same frozen representations.

\begin{table}[t]
\centering
\caption{Main experimental contrasts.
Each row changes a component while preserving the listed parts of the evaluation.}
\label{tabExperimentalContrasts}
\makegapedcells
\begin{tabular}{@{}>{\raggedright\arraybackslash}p{0.3\columnwidth}
>{\raggedright\arraybackslash}p{0.67\columnwidth}@{}}
\toprule
\textbf{Contrast} & \textbf{Quantities Held Fixed}
\\
\midrule
Raw vs.\ One Encoder & Accounts, input features, logistic classifier, folds, and threshold rule
\\
One Fixed vs.\ Three Adaptive & Base features and classifier family, view rule and number of encoder-classifier pairs change
\\
Fixed vs.\ Adaptive Views & Accounts, encoder count, features, and classifier
\\
Raw vs.\ \system RBF & Accounts, input features, kernel classifier, folds, and threshold rule
\\
Feature Removal & Protocol and architecture, one input family is removed
\\
Label Budget & Representation, folds, and paired labeled subset, only label count changes
\\
\bottomrule
\end{tabular}
\end{table}

\subsection{Performance and Label Efficiency}

We pool held-out predictions within each outer repetition and treat accuracy as the primary account-level measure.
Positive-class $F_1$ describes performance on the designated abuse class, while \gls{rocauc} asks whether the ranking improves independently of the chosen threshold \cite{sokolova2009systematic,fawcett2006roc}.
\Cref{secMetricsAndStatisticalSummaries} defines these measures and the paired summaries.

\begin{definition}[Paired Account Contrast]
\label{defPairedAccountContrast}
Let $\mathcal D_A(S)=\{(y_u,\widehat p_{u,A}):u\in S\}$ and $\mathcal D_B(S)$ be the predictions of models $A$ and $B$ on the same held-out account set $S$.
For an account-level metric $M$, their paired contrast is:
$
\Delta_M(S)
=
M\!\left(\mathcal D_A(S)\right)-M\!\left(\mathcal D_B(S)\right)
.
$
\end{definition}

This contrast is the primary estimand for model comparisons because it conditions on the evaluated accounts and their labels.
For the reported accuracy contrasts, a bootstrap resamples accounts and retains the paired correctness outcomes across repetitions \cite{efron1993bootstrap}.
The resulting percentile \glspl{ci} treat the account, rather than the event, as the sampling unit.
Repeated predictions for a sampled account move together in each bootstrap draw, thus preserving their dependence.

The label-efficiency experiment keeps fitted representations and outer splits fixed and varies only the labeled accounts available to the classifier.
Within each training fold, nested stratified subsets contain $1, 2, 5, 10, 25, 50$, or $100\%$ of task labels.
Raw features and \system receive identical labeled accounts at every budget, and pretraining includes every eligible outer-training history but no labels.

\begin{definition}[$\epsilon$-Label Sufficiency]
\label{defEpsilonLabelSufficiency}
Let $\mathcal R$ be the evaluated label fractions, let $A(\rho)$ be mean outer-test accuracy when fraction $\rho\in\mathcal R$ of the outer-training accounts is labeled, and define $\mathcal R_{\ge\rho}=\{\rho'\in\mathcal R:\rho'\ge\rho\}$.
The $\epsilon$-label threshold is:
$$
\begin{aligned}
\rho_\epsilon
=
\min\bigl\{\rho\in\mathcal R:\ 
&\forall\rho'\in\mathcal R_{\ge\rho}, A(\rho')\ge A(1)-\epsilon\bigr\}.
\end{aligned}
$$
\end{definition}

We use $\epsilon=0.01$, so the threshold is the smallest evaluated budget after which accuracy stays within one percentage point of the full-label result.

\section{Results}

\subsection{Does Fission Improve Account Classification?}

\begin{table*}[t]
\centering
\caption{Primary account-level performance.
Positive $F_1$ refers to sockpuppets on Wikipedia and bots on Cresci15.
The Cresci15 rows report repeated outer evaluations: the profile-fallback row covers all 5,301 profiles and the other rows cover the 5,148 accounts with released posts.
A combined Cresci15 ROC-AUC is omitted because its two routing branches are not jointly calibrated.}
\label{tabMainResults}
\resizebox{\textwidth}{!}{
\begin{tabular}{@{}llccc@{}}
\toprule
\textbf{Dataset} & \textbf{Model} & \textbf{Accuracy (\%)} & \textbf{Positive $F_1$ (\%)} & \textbf{ROC-AUC (\%)}
\\
\midrule
Wikipedia Full & \system-Distill & $\mathbf{98.684\pm0.002}$ & $\mathbf{81.957\pm0.025}$ & $\mathbf{94.543\pm0.005}$
\\
\addlinespace
Cresci15 & Raw Features + Logistic & $98.780\pm0.077$ & $99.023\pm0.062$ & $99.787\pm0.034$
\\
& \system, Three Encoders + Logistic & $99.136\pm0.055$ & $99.307\pm0.044$ & $99.921\pm0.012$
\\
& Raw Features + RBF SVM & $98.537\pm0.127$ & $98.827\pm0.101$ & $99.827\pm0.009$
\\
& \system, Three Encoders + RBF SVM & $\mathbf{99.223\pm0.063}$ & $\mathbf{99.377\pm0.051}$ & $\mathbf{99.924\pm0.012}$
\\
& \system{} + Profile Fallback (full cohort) & $\mathbf{99.227\pm0.061}$ & $\mathbf{99.389\pm0.048}$ & ---
\\
\bottomrule
\end{tabular}
}
\end{table*}

The retrospective \system-Distill extension scores all 3,312,087 nonempty Wikipedia account identifiers, reaching $98.684\pm0.002\%$ accuracy, $81.957\pm0.025\%$ sockpuppet $F_1$, and $94.543\pm0.005\%$ \gls{rocauc}.
The one-edit route contains $73.939\%$ of the full cohort.
\Cref{secWikipediaFullCoverage} reports its route-level discrimination and the retrospective protocol boundary.

Within Wikipedia's active cohort, replacing raw inputs with \system representations improves logistic-regression accuracy by $4.037\%$ (paired $95\%$ \gls{ci} $[3.926,4.160]$).
A single fixed-view encoder already improves accuracy by $3.181$ points ($[3.069,3.294]$).

The full three-encoder adaptive model adds a further $0.857$ points ($[0.804,0.910]$).
These controlled active-cohort results explain the teacher representation, while the full-dataset row is the coverage headline.

Cresci15 offers little headroom: the raw logistic model reaches $98.780\%$ accuracy.
The three-encoder representation reaches $99.136\%$ accuracy, a $0.355$-point gain ($[0.190,0.525]$).

With the \gls{rbf} \gls{svm} held fixed, the learned representation adds $0.686$ points, reaching $99.223\%$ compared with $98.537\%$ for the raw inputs.
In contrast to the Wikipedia active cohort, a single fixed-view encoder adds only $0.043$ points, with an interval spanning zero ($[-0.128,0.208]$).
In the matched active-cohort comparisons, \system raises positive $F_1$ and \gls{rocauc} by $4.804$ and $2.298$ points on Wikipedia, and by $0.550$ and $0.097$ points with the Cresci15 \gls{rbf} \gls{svm}.

\subsection{How Many Task Labels Are Needed?}

\begin{figure*}[t]
\centering
\includegraphics[width=0.95\textwidth]{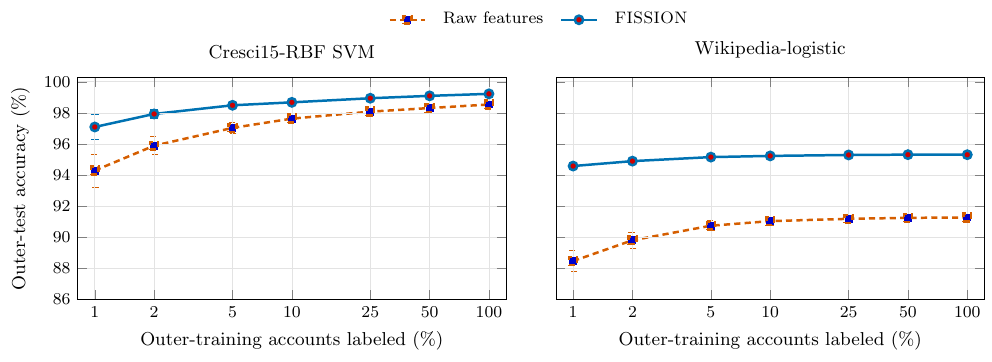}
\caption{Accuracy across label budgets.
Raw features and \system use the same labeled accounts and classifier family per point.
\system pretraining includes every eligible outer-training history but uses none of their labels.
Points show mean pooled out-of-fold accuracy.
Bars show descriptive standard deviations over outer-evaluation/subset combinations.}
\label{figLabelEfficiency}
\end{figure*}

\system's advantage is largest at the smallest label budgets (\cref{figLabelEfficiency}).
In the active-cohort experiment, one percent, or 1,293 labels per fold, gives Wikipedia accuracy of $94.577\%$, $6.114$ points above raw features and $0.729$ points below the full-label result.
On Cresci15, the one-percent advantage is $2.804$ points, while five percent, 206 labels per fold, brings \system within one point of its full-label accuracy.
The resulting $0.01$-label thresholds are $1\%$ for the Wikipedia active cohort and $5\%$ for Cresci15, and \system has higher mean accuracy at every evaluated budget.

\subsection{What Does the Representation Use?}

The ablations point to different signals across platforms.
On the Wikipedia active cohort, removing page-focus and inter-edit-rhythm features lowers accuracy by $10.413$ points, whereas removing lexical edit-comment features lowers it by $0.549$ points.
On Cresci15, removing the lexical dimensions lowers accuracy by $1.946$ points, while the intervals for the behavioral groups include zero.
The component contrasts also differ: on the Wikipedia active cohort, a single fixed-view encoder adds $3.181$ points and adaptive views add $0.299$ in the separate view-rule comparison.
On Cresci15, the corresponding gains are $0.043$ points for the single encoder and $0.128$ or $0.006$ for adaptive views with logistic regression or the \gls{rbf} \gls{svm}.
The contribution of segmentation therefore varies by dataset and classifier.
The principal contrasts and full component, feature, and classifier results appear in \cref{secComponentAndFeatureAblations}.

\section{Discussion}

Temporal fission turns the common origin of two views into supervision before labels are available.
The encoder rewards behavior that recurs within an account.
The labeled sample selects which recurring features predict the task.
The active-cohort ablations reflect this division: page focus and editing rhythm produce the largest Wikipedia loss, whereas text produces the largest Cresci15 loss.
Contiguous, disjoint views make origins informative, because they share no event, the encoder cannot align them through duplication and must rely on distributional regularities that survive the temporal cut.
The relation is weaker than an abuse label and broader than a hand-designed activity proxy.
It identifies observations with a common source while leaving the task classifier to select the useful coordinates.

The benefit is clearest when labels are scarce.
At the one-percent Wikipedia active-cohort budget, the classifier sees 1,293 labeled accounts per training fold while the encoder learns from every eligible training history.
\system then falls only $0.729$ points below its full-label accuracy.
The raw-feature model remains $2.805$ points below its own.
The one-percent Cresci15 fold contains only 41 labeled accounts, and its label-sufficiency threshold is five percent.
These curves show where the gain comes from: pretraining structures the representation before a downstream classifier is fitted on a small labeled sample, while the raw-feature model catches up as more labels become available.
They also describe a two-resource setting.
The unlabeled cohort remains fixed as the task-label budget changes, so each point asks how many abuse judgments are needed once common-history supervision has shaped the encoder.
Raw features and \system receive the same labeled subset, which separates access to account histories from access to their task labels.
The resulting threshold is a measure of label demand conditional on the available histories, not a claim that the same model can be learned from fewer accounts.

Cresci15 is a cross-domain stress test: its raw baseline is nearly saturated, yet \system leads the matched model at every label budget and opens its widest gap at one percent.
Together, the benchmarks show the same principle at work in collaborative editing and social posting.
Each platform supplies its own event summaries.
The common-history relation and training protocol remain unchanged.

The cross-domain ablations argue against a generic activity proxy: editing rhythm carries Wikipedia, while text carries Cresci15.
The common-history relation instead rewards the recurring behavior available on each platform.

The component comparisons also differ by dataset.
One fixed-view encoder explains most of the Wikipedia difference, while Cresci15's change is negligible: the gain appears only when adaptive views and the ensemble are introduced.
As that comparison changes view construction and averaging, it does not assign the gain to either component alone.

Temporal fission defines one invariance: replacing one contiguous period with another should leave the account representation largely unchanged.
The two views are observed behavior, and task labels determine which stable dimensions matter for classification.

The separate view-rule contrasts show that adaptive segmentation has a dataset- and classifier-dependent effect.
Pairing has another tradeoff.
Materializing every within-account pair gives histories with more views greater weight until the 10-view cap, while fixed 10-way fissioning gives every eligible account 45 pairs.
The Wikipedia cohort's raw-to-encoder gain survives the fixed-view comparison, implying common-history supervision is not an artifact of activity-dependent pair counts.
Account-uniform sampling or a multi-positive objective can separate the diversity of temporal comparisons from an account's weight.

As pretraining is class-agnostic, every eligible account contributes before class prevalence is known.
Imbalance enters when the classifier and threshold are fitted.
So, the unlabeled cohort can shape both classes' geometry even when only a small subset is labeled.

\system's contribution likewise lies in the view relation rather than a new contrastive loss.
The \glsentryshort{ntxent} objective and encoder components are standard.
The method turns a complete account history into disjoint, activity-adaptive positives and returns to that complete history for prediction.
The empirical question is whether this construction lowers task-label demand under account-disjoint evaluation.

On Cresci15, the representation adds $0.356$ points with logistic regression and $0.686$ with the \gls{rbf} \gls{svm}, carrying the gain across linear and nonlinear decision boundaries.

The paired analysis is informative near ceiling.
Each contrast compares predictions on the same held-out accounts and bootstraps the difference at the account level.
The $0.686$-point Cresci15 \gls{rbf} \gls{svm} gain therefore measures corrections on a common error set above a $98.537\%$ raw baseline.
Absolute scores describe the resulting detectors and paired intervals describe the change induced by the representation.

The label-sufficiency threshold normalizes each model by its full-label accuracy.
It asks how quickly the model recovers its attainable performance, so the criterion remains meaningful near saturation and when there is more headroom.

Fission requires behavior to persist across several dense windows.
Short histories may not give 2 reliable summaries, while handoffs or abrupt strategy changes can turn nominal positives into mismatched views.
Duration-based or change-point segmentation can condition positives on stable periods, and rolling temporal splits would measure how quickly the learned regularities decay.
Prediction currently embeds the complete history.
Averaging view representations would test the effect of preserving training granularity.
The residual-error analysis locates the same boundary on Cresci15: accounts below the pretraining threshold are overrepresented among persistent errors, though they are still embedded and evaluated from their complete histories.
So, learning useful representations from the first few events is a distinct problem from reducing the number of labels for established histories.

The separation between pretraining and fitting also admits various update rates.
Unlabeled histories can refresh the encoder, while a smaller current sample refits the classifier and its threshold.
Our label-efficiency curves measure the latter requirement within cohorts.

In our experiments, positive pairs are constructed from temporal views of the same account.
If operator identities were available, accounts controlled by the same operator could form positive pairs, with entire operators held out during evaluation.

\section{Conclusion}
An account can supervise its representation before an investigator assigns a task label, as disjoint temporal views already share a known origin.
\system uses that relation to improve classification across markedly different domains while retaining most full-label accuracy when labels are scarce, and the \system-Distill extension transfers the active-history signal to sparse prefixes.

\clearpage\newpage

\printbibliography[heading=bibintoc]

\clearpage\newpage

\appendix

\begin{figure*}[t]
\centering
\resizebox{0.95\textwidth}{!}{
\begin{tikzpicture}[
  font=\small,
  >=Latex,
  stage/.style={
    draw=black!65,
    rounded corners=2pt,
    minimum height=1.15cm,
    align=center,
    fill=black!2,
    inner sep=6pt
  },
  data/.style={stage,fill=blue!7},
  learned/.style={stage,fill=orange!9},
  flow/.style={-{Latex[length=1.7mm,width=1.25mm]},line width=0.65pt,black!75,shorten <=1.1pt,shorten >=1.1pt},
  transfer/.style={-{Latex[length=1.7mm,width=1.25mm]},densely dashed,line width=0.65pt,black!60,shorten <=1.1pt,shorten >=1.1pt},
  transferlabel/.style={
    fill=white,
    inner xsep=2pt,
    inner ysep=1pt,
    font=\footnotesize,
    text=black!70
  }
]
\node[data,minimum width=2.55cm] (history) {
  \textbf{Unlabeled History} $E_u$ \\ [-1pt]
  $\bullet\ \bullet\ \bullet\ \bullet\ \bullet\ \bullet\ \bullet\ \bullet$
};
\node[data,minimum width=3.15cm,right=0.42cm of history] (views) {
  \textbf{Fissioned Temporal Views} \\ [-1pt]
  $\boxed{E_u^{(1)}}\ \boxed{E_u^{(2)}}\ \cdots\ \boxed{E_u^{(m_u)}}$
};
\node[learned,minimum width=2.45cm,right=0.42cm of views] (encoder) {
  \textbf{Shared Encoder} $f_\theta$ \\ [-1pt]
  $h_u^{(j)}=f_\theta(x(E_u^{(j)}))$
};
\node[learned,minimum width=2.45cm,right=0.42cm of encoder] (projector) {
  \textbf{Projection Head} $g_\psi$ \\ [-1pt]
  $z_u^{(j)}=g_\psi(h_u^{(j)})$
};
\node[learned,minimum width=2.85cm,right=0.42cm of projector] (loss) {
  \textbf{NT-Xent Objective} \\ [-1pt]
  Align One Pair from\\the Same History
};
\draw[flow] (history) -- (views);
\draw[flow] (views) -- (encoder);
\draw[flow] (encoder) -- (projector);
\draw[flow] (projector) -- (loss);
\node[learned,minimum width=2.45cm,below=1.48cm of encoder] (embedding) {
  \textbf{Account Representation} \\ [-1pt]
  $h_u=f_\theta(x(E_u))$
};
\node[data,minimum width=3.15cm,left=1.10cm of embedding] (full) {
  \textbf{Complete History} $E_u$
};
\node[learned,minimum width=2.45cm,right=1.10cm of embedding] (classifier) {
  \textbf{Task Classifier} $q_\phi$ \\ [-1pt]
  Fit on Labeled Accounts
};
\node[data,minimum width=2.85cm,right=1.10cm of classifier] (prediction) {
  \textbf{Account Prediction} \\ [-1pt]
  Bot or Human\\Sockpuppet or Benign
};
\draw[flow] (full) -- (embedding);
\draw[flow] (embedding) -- (classifier);
\draw[flow] (classifier) -- (prediction);
\draw[transfer]
  (encoder.south) --
  node[pos=0.5,right=3pt,transferlabel] {Retain $f_\theta$}
  (embedding.north);
\node[
  fit=(history)(views)(encoder)(projector)(loss),
  draw=blue!45,
  rounded corners=3pt,
  inner sep=5pt,
   label={[text=blue!55!black]above:\textbf{Self-Supervised Pretraining}}
] {};
\node[
  fit=(full)(embedding)(classifier)(prediction),
  draw=orange!55,
  rounded corners=3pt,
  inner sep=5pt,
   label={[text=orange!55!black]below:\textbf{Task-Specific Classification}}
] {};
\end{tikzpicture}%
}
\caption{An illustration of the \system pipeline.}
\label{figPipeline}
\end{figure*}
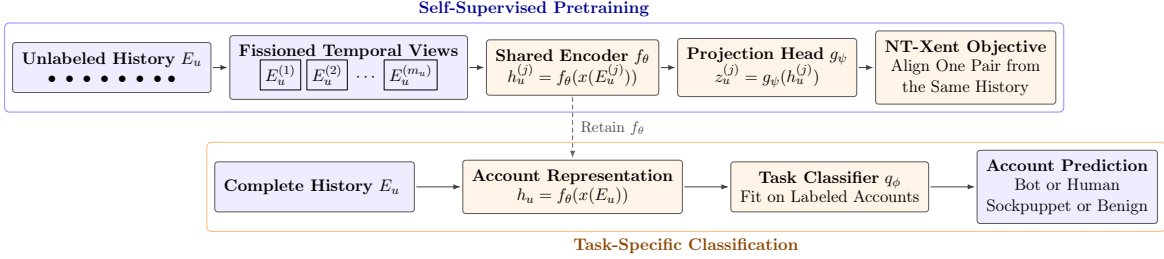

\section{Reproducibility}
\label[appendix]{secReproducibility}

In the appendices, we provide more information on the account-level cohort, feature, training, and evaluation details used in the paper.
Upon the paper's publication, we will open source our pipeline and the aggregates used to produce all tables and figures.
We use only existing public datasets.

\section{Datasets and Cohort Construction}
\label[appendix]{secDatasetsAndCohortConstruction}

\subsection{Cresci15}

The Twitter experiment uses the Cresci15 dataset \cite{cresci2015fame}.
The profile tables contain 5,301 accounts, including 3,351 bots and 1,950 humans.
Of these accounts, 5,148 have released event rows: 3,202 bots and 1,946 humans.
The remaining 153 profiles are evaluated using the fallback described in \cref{secNoPostFallback}.
Within each outer fold, training accounts with at least 20 released events are used to construct fission pairs.
Training and test accounts with shorter histories are embedded using all available events.

The stored retweet indicator in this dataset is uniformly zero, so the feature code reconstructs retweets from the Twitter text prefix ``RT @''.
Because Cresci15's collection sources are closely associated with class, we treat it as an in-cohort comparison.

\subsection{Cresci15 Profiles Without Released Posts}
\label[appendix]{secNoPostFallback}

Raw profile tables contain 5,301 accounts, whereas released post rows cover 5,148.
For each outer fold, the other 153 profiles are assigned without labels or collection-source identifiers and scored by a fixed profile-only logistic student trained on the outer-training accounts' train-side \system probabilities.
The student uses 8 profile count-and-ratio features, no target labels during fitting or threshold selection, and a $\frac{1}{2}$ fixed threshold.
It correctly classifies 152 of 153 fallback accounts.
Combined with \system over all 5,301 profiles, we get $99.227\pm0.061\%$ accuracy and $99.389\pm0.048\%$ bot-$F_1$.

\subsection{Wikipedia}

The Wikipedia experiment uses an English edit stream whose supplied summary contains 3,312,088 account identifiers and labels derived from public sockpuppet investigations \cite{solorio2013case,sakib2022automated,raszewski2025detecting}.
The fields are account identifier, timestamp, page title, recorded edit size, edit comment, and row label.
The loader skips 65 edit rows with blank account identifiers, leaving 3,312,087 nonempty account identifiers.
For the 23,369 identifiers whose rows carry conflicting labels, we apply a positive-wins rule: an account is a sockpuppet if any of its rows is positive.
This yields 132,890 sockpuppets and 3,179,197 benign accounts in the full scoring cohort, whose median history contains one edit.

The main experiments retain the 161,604 accounts with at least ten edits: 66,247 sockpuppets and 95,357 benign accounts.
This active cohort represents $4.9\%$ of the supplied identifiers and has $40.993\%$ sockpuppet prevalence.
The retrospective extension in \cref{secWikipediaFullCoverage} separately scores all 3,312,087 nonempty identifiers while not changing the active-cohort experiment.
As the data lack investigation and master-account identifiers, folds are disjoint by account but cannot be verified as disjoint by investigation or operator.

\begin{table}[t]
\centering
\caption{Account counts.
Positive denotes bot on Cresci15 and sockpuppet on Wikipedia.}
\label{tabAppendixCohorts}
\begin{tabular}{@{}lrrr@{}}
\toprule
\textbf{Cohort} & \textbf{Positive} & \textbf{Negative} & \textbf{Total}
\\
\midrule
Cresci15 Event Cohort & 3,202 & 1,946 & 5,148
\\
Cresci15 Full Profile Cohort & 3,351 & 1,950 & 5,301
\\
Wikipedia Supplied Summary & 116,846 & 3,195,242 & 3,312,088
\\
Wikipedia Full Scoring & 132,890 & 3,179,197 & 3,312,087
\\
Wikipedia Active Cohort & 66,247 & 95,357 & 161,604
\\
\bottomrule
\end{tabular}
\end{table}

\subsection{Outer-Fold Populations}

The main Cresci15 and active-cohort Wikipedia experiments use stratified 5-fold account-level \gls{cv}.
\Cref{tabOuterFoldPopulations} gives realized population ranges.
Accounts, and therefore all their events, remain on one side of each outer split.

\begin{table}[t]
\centering
\caption{Accounts per outer fold.
Total test sizes are 1,029-1,030 for Cresci15 and 32,320-32,321 for Wikipedia.}
\label{tabOuterFoldPopulations}
\begin{tabular}{@{}lrrr@{}}
\toprule
\textbf{Dataset} & \textbf{Train} & \textbf{Positive Test} & \textbf{Negative Test}
\\
\midrule
Cresci15 & 4,118-4,119 & 640-641 & 389-390
\\
Wikipedia & 129,283-129,284 & 13,249-13,250 & 19,071-19,072
\\
\bottomrule
\end{tabular}
\end{table}

\section{Features and Fissioned Views}
\label[appendix]{secFeaturesAndFissionedViews}

\subsection{Behavioral and Lexical Inputs}

Each Cresci15 history is mapped to 39 behavioral dimensions: normalized 24-bin hour-of-day and seven-bin day-of-week histograms, followed by posts per day, retweet rate, hashtags per post, mentions per post, URLs per post, the coefficient of variation of inter-post intervals, mean text length, and text-length standard deviation.
The interval coefficient of variation is clipped at $10^4$, and histories with fewer than two valid timestamps receive a zero behavioral vector.

Wikipedia uses 30 dimensions: normalized 24-bin hour histogram, mean edit size, mean and standard deviation of comment length, edits per unique page, and inter-edit time mean and standard deviation.
Histories with fewer than two valid timestamps receive a 0 behavioral vector.
The full-dataset student and sparse-route experts instead use a fixed 65-dimensional singleton-safe vector.
It contains only account-level behavioral, temporal, count, and hashed lexical summaries available at the routed prefix.

For both datasets, we fit the \gls{tfidf} transform to one lowercase word-unigram document per training account, with at most 5,000 terms and minimum document frequency 5 \cite{salton1988term}.
The transform uses smoothed inverse document frequency and $\ell_2$ row normalization.
A truncated \gls{svd} maps the sparse vectors to 50 lexical dimensions \cite{deerwester1990indexing}.
The fitted transform is applied unchanged to training views, training histories, and outer-test histories.
The encoder receives the resulting vectors directly.
Standardization is reserved for the raw-feature controls.

\begin{table}[t]
\centering
\caption{Feature dimensions.
Wikipedia magnitude contains edit-size and comment-length statistics.
Rhythm contains page focus and inter-edit statistics.}
\label{tabFeatureDimensions}
\begin{tabular}{@{}>{\raggedright\arraybackslash}p{0.175\columnwidth}
>{\raggedright\arraybackslash}p{0.5495\columnwidth}rr@{}}
\toprule
\textbf{Dataset} & \textbf{Behavioral Groups} & \textbf{Lexical} & \textbf{Total}
\\
\midrule
Cresci15 & Hour 24, Day 7, Scalars 8 & 50 & 89
\\
Wikipedia & Hour 24, Magnitude 3, Rhythm 3 & 50 & 80
\\
\bottomrule
\end{tabular}
\end{table}

\subsection{View Construction and Pair Weighting}

After sorting a history by time, the adaptive rule in \cref{eqViewCount} forms contiguous views of near-equal size and materializes every unordered pair.
So, account $u$ contributes $\binom{m_u}{2}$ positive pairs.
Below the 10-view cap, an eligible Cresci15 account contributes between 6 to 36 pairs, and 45 under fixed 10-way fissioning.
View-count ablations jointly vary segment density and the number of pairs contributed by an account.

\begin{figure}[t]
\centering
\includegraphics[width=\columnwidth]{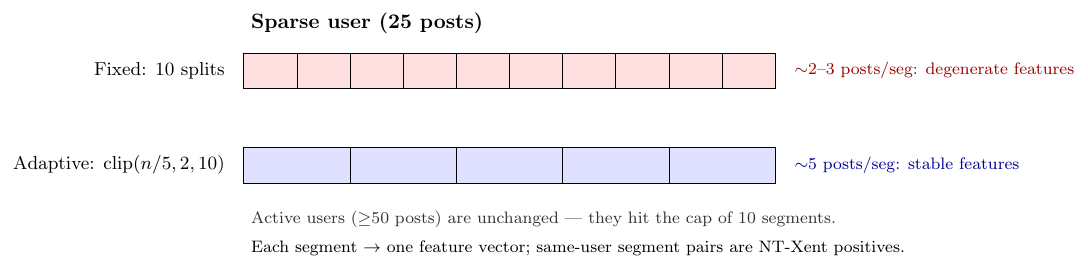}
\caption{Fixed and count-adaptive fissioning of a 25-event history.
The adaptive rule forms five denser views.
Both rules reach ten views at 50 events.}
\label{figSegmentation}
\end{figure}

\subsection{Encoder and Contrastive Objective}

The encoder widths are $d_{\mathrm{in}}\!\rightarrow128\rightarrow128\rightarrow64$, with \gls{bn} and \gls{relu} in the hidden layers and an $\ell_2$-normalized output \cite{ioffe2015batch,nair2010rectified}.
The training-only projector maps $64\!\rightarrow128\!\rightarrow32$ through an intermediate \gls{relu} and normalizes again.
Unit normalization makes each dot product in \cref{eqNtxent} a cosine similarity.

We use single-positive \gls{ntxent} \cite{chen2020simclr}.
Each sampled pair supplies the numerator.
An additional same-account view in the batch remains in the denominator.
Masking these terms changes mean Cresci15 accuracy by $-0.016$ points.
At inference, each complete history is embedded once.
Segment representations are not averaged.

\section{Complete Experimental Protocol}
\label[appendix]{secCompleteExperimentalProtocol}

For the main experiments, data-dependent quantities are recomputed within the training partition of each outer fold, including vocabulary, \gls{svd}, temporal views, encoder, feature standardizer, classifier hyperparameters, probability calibration, and decision thresholds.
A 3-fold inner \gls{cv} chooses classifiers by positive-class $F_1$, then accuracy, then a fixed lower-complexity tie rule.
The final threshold maximizes positive-class $F_1$ on aggregated inner out-of-fold scores.

\begin{table*}[t]
\centering
\caption{Configurations for the main experiments.
Hyperparameters and thresholds are reselected within every outer fold.}
\label{tabExperimentalConfigurations}
\makegapedcells
\begin{tabular}{@{}>{\raggedright\arraybackslash}p{0.15\textwidth}
>{\raggedright\arraybackslash}p{0.4\textwidth}
>{\raggedright\arraybackslash}p{0.4\textwidth}@{}}
\toprule
\textbf{Component} & \textbf{Cresci15} & \textbf{Wikipedia}
\\
\midrule
Eligibility & At least 20 events to form views, all accounts evaluated & Cohort and view eligibility require at least 10 edits
\\
Views & Adaptive $m_u=\operatorname{clip}(\lfloor n_u/5\rfloor,2,10)$, fixed ten-way comparison, all unordered pairs & Adaptive views for the complete model, fixed ten-way views for the single-encoder comparison
\\
Input & 39 behavioral + 50 lexical dimensions & 30 behavioral + 50 lexical dimensions
\\
Text Transform & \multicolumn{2}{p{0.76\textwidth}}{\raggedright Outer-training \gls{tfidf} with 5,000 terms and minimum document frequency 5, followed by a 50-dimensional truncated \gls{svd}.}
\\
Encoder / Projector & \multicolumn{2}{p{0.76\textwidth}}{\raggedright $d_{\mathrm{in}}$-128-128-64 encoder with \gls{bn}, \gls{relu}, and normalized output, training-only 64-128-32 projector with \gls{relu} and normalization.}
\\
Contrastive Fit & \multicolumn{2}{p{0.76\textwidth}}{\raggedright Adam, learning rate $10^{-3}$, 50 epochs, 256 pairs (512 views), temperature $0.12$, single-positive pairs, and diagonal masking.}
\\
Ensemble & Three independent encoders, separate 64-dimensional classifiers, average predicted probabilities & Same for the complete model, one encoder for the core comparison
\\
Outer / Inner CV & Five outer folds repeated $10$ times, three inner folds & Five outer folds repeated $3$ times, three inner folds
\\
Classifier & Per-encoder RBF SVM: $C\in\{1,10,100\}$, $\gamma\in\{\mathrm{scale},0.1\}$, and training-only Platt calibration & Per-encoder logistic regression: $C\in\{0.01,0.1,1,10,100\}$, ordinary or balanced class weights, and at most 1,000 iterations
\\
Matched Classifier & Per-encoder logistic regression with the Wikipedia grid & One tuned logistic model on raw inputs
\\
Threshold & \multicolumn{2}{p{0.76\textwidth}}{\raggedright Selected from aggregated inner out-of-fold scores before outer-test scoring.}
\\
\bottomrule
\end{tabular}
\end{table*}

\subsection{Classifier Details}

The logistic models use an $\ell_2$ penalty, with $C$ denoting inverse regularization strength.
Balanced class weighting assigns classes a weight inversely proportional to training frequency.
Following \citet{cortes1995support}, the \gls{rbf} \gls{svm} uses the kernel $k(h,h')=\exp\!\left(-\gamma\lVert h-h'\rVert_2^2\right)$.
Because encoder outputs are normalized, this kernel is monotone in their cosine similarity.
For the training representation matrix $H_{\mathrm{tr}}$, $\gamma=\mathrm{scale}$ equals $1/(d\,\widehat{\operatorname{Var}}(H_{\mathrm{tr}}))$, with the variance taken over all entries.
Platt calibration is fit to inner out-of-fold decision values before averaging probabilities \cite{platt1999probabilistic}.

Raw-feature controls share the outer accounts, labels, fitted lexical transforms, inner-fold selection, and threshold rule.
Within the relevant training partition, a raw coordinate is standardized as $x_j\mapsto(x_j-\widehat\mu_j)/\widehat\sigma_j$.
With logistic regression fixed, the main comparison measures the value of the complete learned representation.
The single-encoder and component experiments separate its principal design choices.

\subsection{Compute Environment}

Encoder fitting ran on Ubuntu 24.04.4 LTS with an Intel Xeon w5-2445 CPU, 256 GB RAM, and an NVIDIA RTX 4080 with 16 GB memory.
The main environment used Python 3.11.15, NumPy 2.4.6, pandas 3.0.3, scikit-learn 1.9.0, SciPy 1.17.1, joblib 1.5.3, and PyTorch 2.12.1 with CUDA 13.0.
XGBoost 3.2.0 and LightGBM 4.6.0 support the classifier comparison \cite{chen2016xgboost,ke2017lightgbm}.
TabPFN 2.2.1 \cite{hollmann2025tabpfn} ran in an isolated Python 3.11.15 environment with scikit-learn 1.6.1.
Encoder fitting used the GPU, while feature extraction, downstream classification, and analysis used the CPU.

\section{Metrics and Statistical Summaries}
\label[appendix]{secMetricsAndStatisticalSummaries}
\Cref{defBinaryPerformance} uses classic $F$-measure and probability-of-ranking interpretations of \gls{rocauc} \cite{vanrijsbergen1979information,bamber1975area,sokolova2009systematic,fawcett2006roc}.
Cresci15's positive class is bot, Wikipedia's is sockpuppet.
\begin{definition}[Binary Performance]
\label{defBinaryPerformance}
For score $s_u$, threshold $t$, and prediction $\widehat y_u(t)=\mathbf 1[s_u\ge t]$, let $\mathrm{TP}$, $\mathrm{FP}$, $\mathrm{FN}$, $\mathrm{TN}$ be the usual confusion counts, and $S^+$ and $S^-$ be independent scores respectively drawn from the positive and negative classes, and define:
\begin{align*}
    N
    &
    =
    \mathrm{TP}+\mathrm{FP}+\mathrm{FN}+\mathrm{TN}
    ,\\
    \operatorname{Acc}
    &
    =
    \frac{\mathrm{TP}+\mathrm{TN}}{N}
    ,\\
    P
    &
    =
    \frac{\mathrm{TP}}{\mathrm{TP}+\mathrm{FP}}
    ,\qquad
    R
    =
    \frac{\mathrm{TP}}{\mathrm{TP}+\mathrm{FN}}
    ,\\
    F_1
    &
    =
    \frac{2\mathrm{TP}}{2\mathrm{TP}+\mathrm{FP}+\mathrm{FN}}
    ,\\
    \operatorname{BAcc}
    &
    =
    \frac{1}{2}
    \left(
    \frac{\mathrm{TP}}{\mathrm{TP}+\mathrm{FN}}
    +
    \frac{\mathrm{TN}}{\mathrm{TN}+\mathrm{FP}}
    \right)
    ,\\
    \operatorname{MacroF}_1
    &
    =
    \frac{F_1^{(+)}+F_1^{(-)}}{2}
    ,\\
    \operatorname{AUC}
    &
    =
    \Pr(S^+>S^-)+\tfrac12\Pr(S^+=S^-)
    .
\end{align*}
\end{definition}

For the full-dataset, \gls{ap} is computed with \texttt{sklearn.metrics.average\_precision\_score}.

\begin{definition}[Pooled Out-of-Fold Summary]
\label{defPooledOutOfFoldSummary}
For repetition $r$, concatenate the five outer-test folds so that every account contributes one tuple $(y_u,s_{ur},\widehat y_{ur})$, and compute metric $M_r$ on this pooled set.
For $R$ repetitions, report
\begin{align*} 
\overline M
&
=
\frac{1}{R}\sum_{r=1}^R M_r,
\\
s_M
&
=
\sqrt{\frac{1}{R-1}\sum_{r=1}^R(M_r-\overline M)^2}.
\end{align*}
\end{definition}

Pooling predictions before evaluating $M_r$ weights every account equally and avoids averaging slightly different sized folds, and $s_M$ describes variation across outer repetitions.

\begin{definition}[Paired Account Bootstrap for Accuracy]
\label{defPairedAccountBootstrapForAccuracy}
For models $A$ and $B$, let $c_{uAr}=\mathbf 1[\widehat y_{uAr}=y_u]$ denote the correctness of model $A$ for account $u$ in repetition $r$, and define $d_u=\frac{1}{R}\sum_{r=1}^{R}(c_{uAr}-c_{uBr})$.
For $N$ evaluated accounts, draw $u_1^{*(b)},\ldots,u_N^{*(b)}$ with replacement and set $\Delta^{*(b)}=\frac{1}{N}\sum_{\ell=1}^{N}d_{u_\ell^{*(b)}}$.
The $95\%$ interval is given by the $2.5$th and $97.5$th percentiles of $\{\Delta^{*(b)}\}_{b=1}^{B_{\mathrm{boot}}}$.
\end{definition}
The bootstrap resamples account-level differences $d_u$ with replacement, so outcomes for models $A$ and $B$ remain paired for each account \cite{efron1993bootstrap,field2007bootstrapping}.
For $\widehat{\Delta}=N^{-1}\sum_u d_u$, the two-sided centered-bootstrap value reported in \cref{tabCresci15FeatureGroups} is:
$$
p_{\mathrm{boot}}
=
\frac{1+\sum_{b=1}^{B_{\mathrm{boot}}}
\mathbf 1\!\left[
\left|\Delta^{*(b)}-\widehat{\Delta}\right|
\ge
\left|\widehat{\Delta}\right|
\right]}
{B_{\mathrm{boot}}+1}.
$$
Centering gives an approximate bootstrap null, and adding $1$ prevents a 0 Monte Carlo value \cite{hall1991guidelines,phipson2010permutation}.
Families of related feature-removal tests use \citet{holm1979simple}'s sequential adjustment.

\section{Retrospective Wikipedia Full-Dataset Coverage}
\label[appendix]{secWikipediaFullCoverage}

This extension transfers the frozen active-account \system signal to histories too short for the original ten-edit evaluation.
For each teacher seed, the frozen outer-test probabilities supervise a pooled sparse student without using sparse-account task labels.
One-edit accounts use a logistic student with median imputation, robust scaling, and $\ell_2$ regularization ($C=1$, seed 20260729).
Accounts with two through nine edits use fixed XGBoost regressors, one per edit-count route, to predict the mean frozen \system probability for the corresponding chronological prefix \cite{chen2016xgboost}.
Each expert uses 200 CPU histogram trees, depth four, learning rate $0.05$, minimum child weight 100, and unit row and column subsampling, $\lambda=1$, $\alpha=0$.
Accounts with at least ten edits retain their seed-specific frozen \system score.
Student and expert outputs are generated and sealed before sparse-account task labels are read.

All views from an account are assigned to the same one of 5 deterministic account-key folds.
Within each route, held-out scores receive clipped-logit Platt calibration fitted on the other 4 folds.
These folds determine a global threshold that maximizes positive-class $F_1$, and the threshold is then applied to the held-out fold.
Thus, every account is excluded from its own calibration and threshold fit, although target labels from other folds are used in both steps.
For each teacher seed, metrics are computed after pooling the 5 held-out folds.
We provide mean and sample standard deviation in \cref{tabWikipediaFullCoverage}.

Positive-class $F_1$ is the headline operating-point objective.
The full-row $F_{0.5}$ is computed separately for each seed from its sealed $\mathrm{TP}$, $\mathrm{FP}$, and $\mathrm{FN}$ counts before taking the mean and sample standard deviation.
The additional full-dataset metrics omitted from the main-text are $87.104\pm0.012\%$ balanced accuracy and $90.637\pm0.013\%$ macro-$F_1$.
This output has the highest sockpuppet $F_1$, accuracy, macro-$F_1$, and sockpuppet precision among the evaluated final outputs.

The one-edit route is the main bottleneck: it contains 2,448,931 accounts ($73.939\%$ of the full cohort) and obtains $68.228\%$ ROC-AUC and $5.962\%$ AP.
The two-edit route obtains $93.726\%$ ROC-AUC and $66.017\%$ AP.
For the three- through nine-edit routes, ROC-AUC ranges from $97.046\%$ to $98.997\%$, and AP ranges from $86.257\%$ to $97.264\%$.
The score-before-label seal verifies that sparse student and expert scores were produced without sparse-account task labels.

\begin{table*}[t]
\centering
\caption{Wikipedia results in the column order of the main-text comparison in \cref{tabCompareSockpuppet}.
The RD25 Reptile baseline and the full-dataset \system-Distill result are repeated from the main-text table.
The 161,604-account active-cohort row shows the teacher model from which the sparse-history students are distilled.
RD25 reports AUPRC, the active row uses trapezoidal AUPRC, and the full row uses Average Precision (AP).
}
\label{tabWikipediaFullCoverage}
\resizebox{\textwidth}{!}{
\begin{tabular}{lccccccc}
\toprule
\textbf{Approach} & \textbf{AUROC} & \textbf{AUPRC / AP} & \textbf{F1-Score} & \textbf{F0.5-Score} & \textbf{Accuracy} & \textbf{Precision} & \textbf{Recall}
\\
\midrule
Reptile Enc. [RD25] & $78.98\pm0.12$ & $62.21\pm0.08$ & $67.46\pm0.53$ & $67.89\pm0.17$ & $77.51\pm0.19$ & $69.43\pm0.26$ & $70.81\pm0.82$
\\
\midrule
\textbf{Our Work (Active)} & $98.917\pm0.013$ & $98.437\pm0.013$ & $94.307\pm0.014$ & $93.992\pm0.030$ & $95.306\pm0.013$ & $93.783\pm0.044$ & $94.838\pm0.039$
\\
\textbf{Our Work (Full)} & $94.543\pm0.005$ & $84.084\pm0.012$ & $81.957\pm0.025$ & $87.182\pm0.057$ & $98.684\pm0.002$ & $91.052\pm0.084$ & $74.515\pm0.027$
\\
\bottomrule
\end{tabular}}
\end{table*}

\section{Component and Feature Ablations}
\label[appendix]{secComponentAndFeatureAblations}

\begin{table*}
\centering
\caption{Principal architecture and feature contrasts in accuracy percentage points.
The Wikipedia view-rule comparison uses a separate 30,000-account cohort, while all other rows use the main evaluated cohorts.}
\label{tabComponents}
\begin{tabular}{@{}>{\raggedright\arraybackslash}p{0.1\textwidth}
>{\raggedright\arraybackslash}p{0.225\textwidth}rr@{}}
\toprule
\textbf{Dataset} & \textbf{Contrast} & \textbf{$\Delta$Acc.} & \textbf{95\% CI}
\\
\midrule
Wikipedia & Raw $\rightarrow$ One Encoder & $+3.181$ & $[3.069,3.294]$
\\
& One $\rightarrow$ Three Adaptive & $+0.857$ & $[0.804,0.910]$
\\
& Fixed $\rightarrow$ Adaptive Views & $+0.299$ & $[0.188,0.411]$
\\
& Remove Rhythm & $-10.413$ & $[-10.583,-10.245]$
\\
& Remove Text & $-0.549$ & $[-0.619,-0.477]$
\\
\addlinespace
Cresci15 & Raw $\rightarrow$ One Encoder & $+0.043$ & $[-0.128,0.208]$
\\
& Fixed $\rightarrow$ Adaptive, Logistic & $+0.128$ & $[0.049,0.210]$
\\
& Fixed $\rightarrow$ Adaptive, RBF & $+0.006$ & $[-0.047,0.060]$
\\
& Remove Text & $-1.946$ & $[-2.323,-1.585]$
\\
\bottomrule
\end{tabular}
\end{table*}

\subsection{Count-Adaptive Views and Classifier}

\Cref{tabFissionClassifierFactorial} holds the three encoders, folds, and features fixed while crossing temporal fissioning with the classifier family.
Adaptive views help the logistic model more than the \gls{rbf} \gls{svm}, while the kernel contributes more under fixed views.

\begin{table}[t]
\centering
\caption{Cresci15 fissioning-by-classifier comparison.
Adaptive minus fixed accuracy is $+0.128$ points with logistic regression (95\% CI $[+0.049,+0.210]$) and $+0.006$ with the RBF SVM ($[-0.047,+0.060]$).}
\label{tabFissionClassifierFactorial}
\begin{tabular}{@{}llrr@{}}
\toprule
\textbf{Views} & \textbf{Classifier} & \textbf{Accuracy (\%)} & \textbf{Bot $F_1$ (\%)}
\\
\midrule
Fixed Ten-Way & Logistic & $99.007\pm0.050$ & $99.205\pm0.040$
\\
Count-Adaptive & Logistic & $99.136\pm0.055$ & $99.307\pm0.044$
\\
Fixed Ten-Way & RBF SVM & $99.217\pm0.061$ & $99.372\pm0.049$
\\
Count-Adaptive & RBF SVM & $\mathbf{99.223\pm0.063}$ & $\mathbf{99.377\pm0.051}$
\\
\bottomrule
\end{tabular}
\end{table}

Wikipedia uses a frozen 30,000-account sample drawn from the active cohort in proportion to label and 4 edit-count strata: 10-19, 20-99, 100-499, and at least 500 edits.
With cohort seed fixed independently of model fitting, each of 3 outer repetitions compares fixed and adaptive 3-encoder models on the same 5 folds.
On this cohort, count-adaptive views improve accuracy by $0.299$ points ($95\%$ \gls{ci} $[0.188,0.411]$) relative to fixed 10-way views.

\subsection{Cresci15 Feature Groups}

The leave-1-group-out experiment preserves encoder width and replaces 1 input group with zeros.
Removing text lowers accuracy by $1.9$ points.
The other removal intervals include 0.

\begin{table}[t]
\centering
\caption{Cresci15 leave-one-group-out feature analysis.
The all-feature row is the separately fitted paired baseline for this ablation.
The Holm family contains the four removal variants, and text-only is an additional reference point.}
\label{tabCresci15FeatureGroups}
\begin{tabular}{@{}lrrrr@{}}
\toprule
\textbf{Variant} & \textbf{Acc. (\%)} & \textbf{$\Delta$ (pp)} & \textbf{95\% CI} & \textbf{Holm $p$}
\\
\midrule
All Features & $99.242\pm0.066$ & - & - & -
\\
Without Hour & 99.211 & $-0.031$ & $[-0.120,+0.060]$ & 0.73
\\
Without Day & 99.180 & $-0.062$ & $[-0.150,+0.020]$ & 0.50
\\
Without Scalars & 99.176 & $-0.066$ & $[-0.210,+0.070]$ & 0.73
\\
Without Text & 97.296 & $\mathbf{-1.946}$ & $\mathbf{[-2.323,-1.585]}$ & $\mathbf{0.0004}$
\\
Text Only & 99.083 & $-0.159$ & $[-0.310,-0.020]$ & -
\\
\bottomrule
\end{tabular}
\end{table}

\subsection{Wikipedia Feature Blocks}

The full-scale Wikipedia analysis retrains the complete model after removing one input block.
Removing page focus and inter-edit rhythm produces the largest observed loss while removing lexical edit-comment features produces a smaller but consistent loss.

\begin{table}[t]
\centering
\caption{Wikipedia feature-block analysis.
The all-feature row is the separately fitted paired baseline for this ablation.
Negative changes denote lower accuracy.}
\label{tabWikipediaFeatureBlocks}
\begin{tabular}{@{}lrrr@{}}
\toprule
\textbf{Variant} & \textbf{Accuracy (\%)} & \textbf{Change (pp)} & \textbf{95\% CI (pp)}
\\
\midrule
All Features & 95.314 & - & -
\\
Without Rhythm & 84.902 & $-10.413$ & $[-10.583,-10.245]$
\\
Without Text & 94.765 & $-0.549$ & $[-0.619,-0.477]$
\\
Text Only & 77.991 & $-17.323$ & $[-17.516,-17.129]$
\\
\bottomrule
\end{tabular}
\end{table}

\section{Classifier-Family Robustness}
\label[appendix]{secClassifierFamilyRobustness}
TabPFN and \gls{rbf} \gls{svm} tie in \cref{tabClassifierResults}.
We retain \gls{rbf} \gls{svm} as it requires no extra pretrained tabular model and supports CPU inference.
The comparison spans linear log-odds \cite{cox1958regression}, nonlinear maximum-margin classification \cite{cortes1995support}, bagged and boosted trees \cite{breiman2001random,chen2016xgboost,ke2017lightgbm}, cosine-neighbor voting \cite{cover1967nearest}, a shallow \gls{mlp}, and TabPFN's pretrained tabular prior \cite{hollmann2025tabpfn}.
Each family is evaluated by averaging 3 per-encoder probabilities and, where applicable, by fitting one model to the concatenated 192-dimensional representation.
Hyperparameters use the same inner folds, selection metric, and inner out-of-fold threshold.
Logistic regression and shallow \gls{mlp} use the \gls{lmbfgs} quasi-Newton method \cite{liu1989limited}, with both fits capped at 1,000 iterations.

\begin{table}[t]
\centering
\caption{Classifier-family settings.
Tree row and column subsampling equal one unless listed otherwise.}
\label{tabClassifierSettings}
\makegapedcells
\scalebox{1}{\begin{tabular}{@{}>{\raggedright\arraybackslash}p{0.3\columnwidth}
>{\raggedright\arraybackslash}p{0.675\columnwidth}@{}}
\toprule
\textbf{Classifier} & \textbf{Search Grid and Fixed Settings}
\\
\midrule
Logistic Regression & $C\in\{0.01,0.1,1,10,100\}$, ordinary or balanced class weights, \glsentryshort{lmbfgs}, at most 1,000 iterations
\\
RBF SVM & $C\in\{1,10,100\}$, $\gamma\in\{\mathrm{scale},0.1\}$, Platt map fit to inner out-of-fold decision values
\\
XGBoost & trees $\in\{200,500\}$, depth $\in\{2,3,4\}$, learning rate $\in\{0.05,0.1\}$
\\
LightGBM & trees $\in\{200,500\}$, depth $\in\{2,3,4\}$, learning rate $\in\{0.05,0.1\}$
\\
MLP & one hidden layer of width 32 or 128, $\alpha\in\{10^{-4},10^{-2}\}$, \glsentryshort{lmbfgs}, at most 1,000 iterations
\\
Random Forest & 500 trees, maximum features $\sqrt d$ or $0.5d$
\\
Cosine $k$-Nearest Neighbors (\glsentryshort{knn}) & neighbors $\in\{15,51,101\}$, distance weighting, cosine metric
\\
TabPFN & Version 2.2.1 default estimator
\\
\bottomrule
\end{tabular}}
\end{table}

\begin{table}[t]
\centering
\caption{Cresci15 classifier comparison on fixed representations, separate from the repeated end-to-end evaluation in \cref{tabMainResults}.
Accuracy differences are measured against logistic regression on the concatenated representation.}
\label{tabClassifierResults}
\scalebox{0.95}{\begin{tabular}{@{}lrr@{}}
\toprule
\textbf{Classifier and Layout} & \textbf{Accuracy (\%)} & \textbf{$\Delta$Acc. (pp)}
\\
\midrule
TabPFN, Per Encoder & 99.219 & $+0.062$
\\
RBF SVM, Per Encoder & 99.219 & $+0.062$
\\
TabPFN, Joint & 99.207 & $+0.050$
\\
RBF SVM, Joint & 99.200 & $+0.043$
\\
XGBoost, Per Encoder & 99.184 & $+0.027$
\\
Logistic, Joint (Reference) & 99.157 & $0$
\\
LightGBM, Per Encoder & 99.122 & $-0.035$
\\
Random Forest, Per Encoder & 99.099 & $-0.058$
\\
XGBoost, Joint & 99.075 & $-0.082$
\\
MLP, Per Encoder & 99.052 & $-0.105$
\\
MLP, Joint & 98.978 & $-0.179$
\\
Cosine kNN, Per Encoder & 98.936 & $-0.221$
\\
\bottomrule
\end{tabular}}
\end{table}

\section{Complete Label-Efficiency Results}
\label[appendix]{secCompleteLabelEfficiencyResults}

\begin{figure*}[t]
\centering
\includegraphics[width=0.95\textwidth]{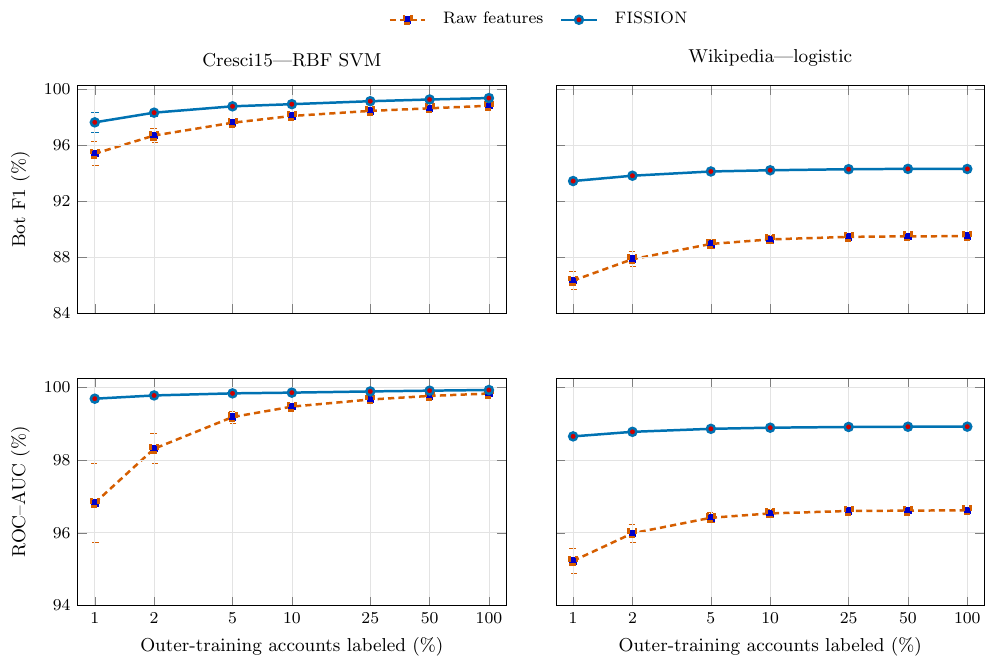}
\caption{Positive-class $F_1$ (top) and ROC-AUC (bottom) across label budgets for Cresci15 (left) and Wikipedia (right).
Orange dotted lines denote raw features and blue solid lines denote \system, while bars show sample standard deviations.}
\label{figLabelEfficiencyMetrics}
\end{figure*}

Label budgets are nested within each outer training fold.
At every budget, raw features and \system use identical stratified subsets.
Pretraining includes all eligible outer-training history but no labels.
Below $100\%$, entries are means and descriptive standard deviations over outer-evaluation/subset combinations.
Full-budget entries summarize outer evaluations.
\Cref{figLabelEfficiencyMetrics} trace positive-class $F_1$ and \gls{rocauc} curves behind \cref{figLabelEfficiency}'s accuracy summary.
At the one-percent budget, \system improves accuracy, positive-class $F_1$, and \gls{rocauc} by $2.804$, $2.239$, and $2.867$ points on Cresci15.
The corresponding gains on Wikipedia are $6.114$, $7.114$, and $3.423$ points.
\Cref{tabCresci15SvmLabelEfficiency,tabWikipediaLabelEfficiency} give budget-by-budget values for the main Cresci15 and Wikipedia classifiers.
\Cref{tabCresci15LogisticLabelEfficiency} gives Cresci15's matched logistic-regression comparison.

\begin{table*}[t]
\centering
\caption{Complete Cresci15 label-efficiency results with the RBF SVM.}
\label{tabCresci15SvmLabelEfficiency}
\setlength{\tabcolsep}{2.5pt}
\begin{tabular}{@{}rrlccc@{}}
\toprule
\textbf{Budget} & \textbf{Labels per Fold} & \textbf{Input} & \textbf{Accuracy (\%)} & \textbf{Positive $F_1$ (\%)} & \textbf{ROC-AUC (\%)}
\\
\midrule
1\% & 41 & Raw & $94.287\pm1.056$ & $95.402\pm0.872$ & $96.819\pm1.087$
\\
&  & \system & $97.091\pm0.817$ & $97.640\pm0.689$ & $99.685\pm0.076$
\\
\addlinespace[1.5pt]
2\% & 82 & Raw & $95.887\pm0.585$ & $96.698\pm0.476$ & $98.309\pm0.414$
\\
&  & \system & $97.929\pm0.318$ & $98.332\pm0.263$ & $99.775\pm0.042$
\\
\addlinespace[1.5pt]
5\% & 206 & Raw & $97.032\pm0.327$ & $97.614\pm0.269$ & $99.181\pm0.164$
\\
&  & \system & $98.487\pm0.163$ & $98.786\pm0.132$ & $99.832\pm0.027$
\\
\addlinespace[1.5pt]
10\% & 412 & Raw & $97.629\pm0.187$ & $98.097\pm0.154$ & $99.467\pm0.060$
\\
&  & \system & $98.676\pm0.139$ & $98.939\pm0.112$ & $99.852\pm0.025$
\\
\addlinespace[1.5pt]
25\% & 1,030 & Raw & $98.078\pm0.130$ & $98.458\pm0.105$ & $99.663\pm0.031$
\\
&  & \system & $98.941\pm0.094$ & $99.151\pm0.075$ & $99.885\pm0.025$
\\
\addlinespace[1.5pt]
50\% & 2,059-2,060 & Raw & $98.310\pm0.111$ & $98.644\pm0.089$ & $99.761\pm0.025$
\\
&  & \system & $99.094\pm0.093$ & $99.273\pm0.075$ & $99.906\pm0.020$
\\
\addlinespace[1.5pt]
100\% & 4,118-4,119 & Raw & $98.537\pm0.127$ & $98.827\pm0.101$ & $99.827\pm0.009$
\\
&  & \system & $99.223\pm0.063$ & $99.377\pm0.051$ & $99.924\pm0.012$
\\
\bottomrule
\end{tabular}
\end{table*}

\begin{table*}[t]
\centering
\caption{Complete Wikipedia label-efficiency results with logistic regression.}
\label{tabWikipediaLabelEfficiency}
\setlength{\tabcolsep}{2.5pt}
\begin{tabular}{@{}rrlccc@{}}
\toprule
\textbf{Budget} & \textbf{Labels per Fold} & \textbf{Input} & \textbf{Accuracy (\%)} & \textbf{Positive $F_1$ (\%)} & \textbf{ROC-AUC (\%)}
\\
\midrule
1\% & 1,293 & Raw & $88.464\pm0.657$ & $86.332\pm0.645$ & $95.226\pm0.345$
\\
&  & \system & $94.577\pm0.098$ & $93.445\pm0.111$ & $98.649\pm0.024$
\\
\addlinespace[1.5pt]
2\% & 2,586 & Raw & $89.801\pm0.529$ & $87.861\pm0.522$ & $95.985\pm0.239$
\\
&  & \system & $94.895\pm0.079$ & $93.826\pm0.074$ & $98.774\pm0.020$
\\
\addlinespace[1.5pt]
5\% & 6,464 & Raw & $90.738\pm0.317$ & $88.949\pm0.323$ & $96.412\pm0.136$
\\
&  & \system & $95.153\pm0.046$ & $94.129\pm0.052$ & $98.857\pm0.016$
\\
\addlinespace[1.5pt]
10\% & 12,928 & Raw & $91.033\pm0.263$ & $89.275\pm0.273$ & $96.531\pm0.121$
\\
&  & \system & $95.230\pm0.035$ & $94.216\pm0.040$ & $98.888\pm0.014$
\\
\addlinespace[1.5pt]
25\% & 32,321 & Raw & $91.182\pm0.145$ & $89.455\pm0.158$ & $96.599\pm0.076$
\\
&  & \system & $95.290\pm0.021$ & $94.291\pm0.028$ & $98.908\pm0.012$
\\
\addlinespace[1.5pt]
50\% & 64,642 & Raw & $91.243\pm0.090$ & $89.495\pm0.091$ & $96.608\pm0.052$
\\
&  & \system & $95.310\pm0.011$ & $94.315\pm0.015$ & $98.915\pm0.011$
\\
\addlinespace[1.5pt]
100\% & 129,283-129,284 & Raw & $91.269\pm0.012$ & $89.503\pm0.008$ & $96.619\pm0.014$
\\
&  & \system & $95.306\pm0.013$ & $94.307\pm0.014$ & $98.917\pm0.013$
\\
\bottomrule
\end{tabular}
\end{table*}

\begin{table*}[t]
\centering
\caption{Cresci15 label efficiency with logistic regression.}
\label{tabCresci15LogisticLabelEfficiency}
\setlength{\tabcolsep}{2.5pt}
\begin{tabular}{@{}rrlccc@{}}
\toprule
\textbf{Budget} & \textbf{Labels per Fold} & \textbf{Input} & \textbf{Accuracy (\%)} & \textbf{Positive $F_1$ (\%)} & \textbf{ROC-AUC (\%)}
\\
\midrule
1\% & 41 & Raw & $93.118\pm0.823$ & $94.352\pm0.711$ & $96.294\pm0.736$
\\
&  & \system & $96.646\pm0.782$ & $97.274\pm0.664$ & $97.720\pm1.932$
\\
\addlinespace[1.5pt]
2\% & 82 & Raw & $95.141\pm0.493$ & $96.061\pm0.415$ & $97.748\pm0.466$
\\
&  & \system & $97.563\pm0.513$ & $98.035\pm0.422$ & $98.370\pm2.298$
\\
\addlinespace[1.5pt]
5\% & 206 & Raw & $96.646\pm0.320$ & $97.298\pm0.258$ & $98.826\pm0.248$
\\
&  & \system & $98.347\pm0.236$ & $98.674\pm0.191$ & $99.668\pm0.104$
\\
\addlinespace[1.5pt]
10\% & 412 & Raw & $97.298\pm0.238$ & $97.829\pm0.193$ & $99.152\pm0.234$
\\
&  & \system & $98.679\pm0.135$ & $98.941\pm0.109$ & $99.813\pm0.064$
\\
\addlinespace[1.5pt]
25\% & 1,030 & Raw & $97.923\pm0.161$ & $98.332\pm0.130$ & $99.538\pm0.151$
\\
&  & \system & $98.930\pm0.102$ & $99.142\pm0.082$ & $99.877\pm0.035$
\\
\addlinespace[1.5pt]
50\% & 2,059-2,060 & Raw & $98.273\pm0.115$ & $98.615\pm0.093$ & $99.678\pm0.070$
\\
&  & \system & $99.079\pm0.086$ & $99.262\pm0.069$ & $99.905\pm0.014$
\\
\addlinespace[1.5pt]
100\% & 4,118-4,119 & Raw & $98.780\pm0.077$ & $99.023\pm0.062$ & $99.787\pm0.034$
\\
&  & \system & $99.136\pm0.055$ & $99.307\pm0.044$ & $99.921\pm0.012$
\\
\bottomrule
\end{tabular}
\end{table*}

\section{Error and Representation Diagnostics}
\label[appendix]{secErrorAndRepresentationDiagnostics}

\subsection{Confusion Matrices}
\begin{figure*}[!t]
\centering
\includegraphics[width=0.95\textwidth]{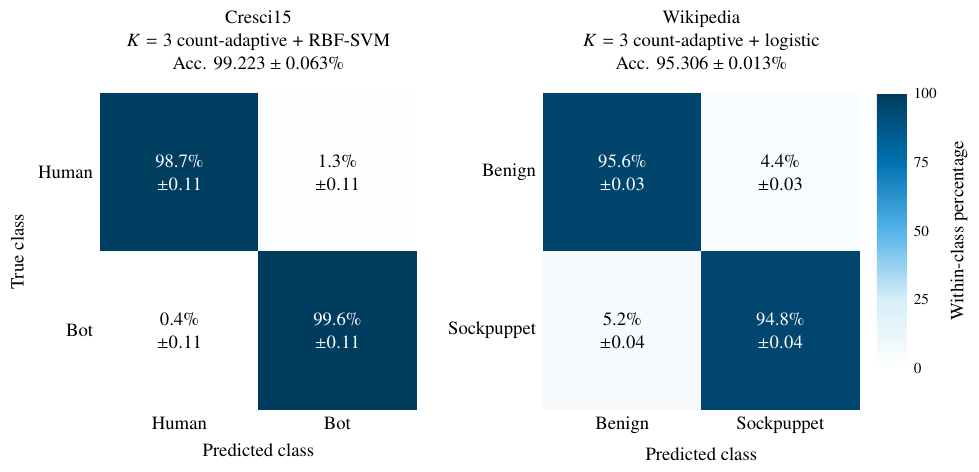}
\caption{Row-normalized out-of-fold confusion matrices for Cresci15 (left) and Wikipedia (right), averaged over the main repeated evaluations.
Cells show mean within-class percentages with sample standard deviations.}
\label{figConfusionMatrices}
\end{figure*}

The classwise error patterns in \cref{figConfusionMatrices} show that the strong aggregate scores are not driven by one class alone.
Recall remains high for both classes: $98.66\%$ for humans and $99.56\%$ for bots on Cresci15, and $95.63\%$ for benign Wikipedia accounts and $94.83\%$ for sockpuppets.

\subsection{Cresci15 Residual-Error Activity}
\begin{figure}
\centering
\includegraphics[width=0.6\columnwidth]{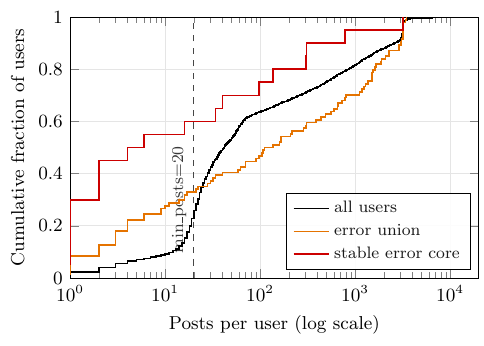}
\caption{Event-count \glspl{ecdf} for all Cresci15 accounts, accounts misclassified at least once, and accounts misclassified in every repeated evaluation.
The dashed line marks the 20-event contrastive-eligibility threshold.}
\label{figECDF}
\end{figure}

Aggregate recall does not reveal how the remaining errors relate to history length, so \cref{figECDF} compares activity across the full Cresci15 cohort and the two error subsets.

Across repeated evaluations, 94 Cresci15 accounts are misclassified at least once and 20 every time.
Sixty percent of the consistently misclassified accounts have fewer than 20 events, compared with $33\%$ of the ever-misclassified accounts and $23\%$ of the cohort.
Sparse histories are thus overrepresented among persistent errors, although activity count is intertwined with class and collection source.

\subsection{Directly Computed Feature Contrasts}

Cliff's rank-based effect size \cite{cliff1993dominance} summarizes univariate class separation.
Cliff's $\delta$ ranges from $-1$ to $1$, with its sign indicating which class tends to have larger values.
Cresci15 classes differ on several activity and interaction statistics.
Wikipedia's largest differences are in edits per page and inter-edit timing, consistent with feature-removal results.

\begin{definition}[Cliff's $\delta$ \cite{cliff1993dominance}]
\label{defCliffsDelta}
Let $X^+\sim F_+$, $X^-\sim F_-$ be independent draws, then $\delta$ is: $$\delta(X^+,X^-)=\Pr(X^+>X^-)-\Pr(X^+<X^-).$$
For samples $x_1^+,\ldots,x_{n_+}^+$ and $x_1^-,\ldots,x_{n_-}^-$, its estimator is:
$$
\widehat\delta
=
\frac{1}{n_+n_-}\sum_{i=1}^{n_+}\sum_{j=1}^{n_-}\operatorname{sgn}(x_i^+-x_j^-)
.
$$
\end{definition}

\begin{figure*}[t]
\centering
\includegraphics[width=0.95\textwidth]{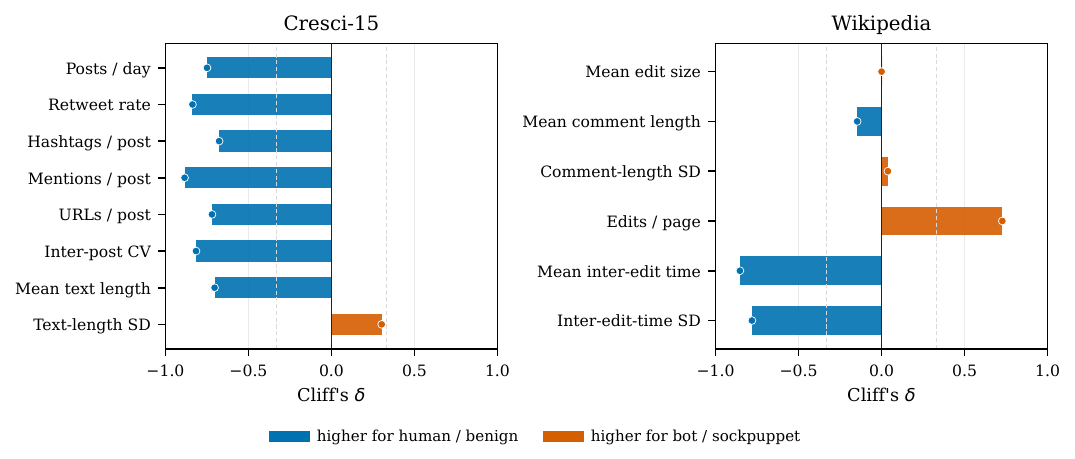}
\caption{Cliff's $\delta$ for directly computed scalar features on balanced samples of 1,000 accounts per class: Cresci15 (left) and Wikipedia (right).
Positive values indicate larger values for bots or sockpuppets.
Fitted lexical coordinates are omitted.}
\label{figFeatureEffectSizes}
\end{figure*}

\subsection{Input and Representation Geometry}

\begin{figure*}[t]
\centering
\includegraphics[width=0.8\textwidth]{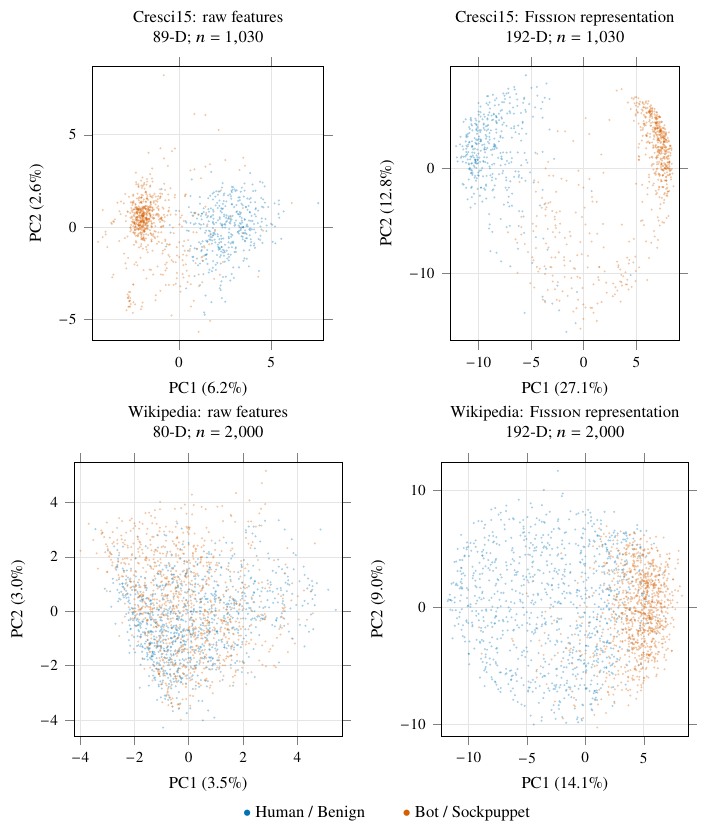}
\caption{Two-dimensional \glsentryshort{pca} summaries \cite{jolliffe2016pca} of raw inputs and concatenated three-encoder representations in one prespecified outer-test fold per dataset: the first fold of the first reported repetition, containing 1,030 Cresci15 accounts and a class-balanced sample of 2,000 Wikipedia accounts.
Each projection is fitted separately on the corresponding training features and applied to the displayed test accounts.}
\label{figFeaturePCA}
\end{figure*}

\Cref{figFeaturePCA} uses \gls{pca} to project centered observations onto the leading eigenvectors of their sample covariance matrix.
Both datasets retain class overlap in two dimensions, most visibly for Wikipedia raw features.

\section{Scope}
The available data identify accounts but not common operators or investigations.
Wikipedia therefore permits account-disjoint but not operator-disjoint folds, while Cresci15 retains the class-associated source structure of its benchmark.
The reported estimates describe retrospective performance within these cohorts.
The full-dataset extension is additionally target-label-aware at route calibration and threshold selection: each account is excluded from its own fitted calibrator and threshold, but labels from other folds are used.
Its pooled ROC-AUC and AP characterize the calibrated routed score and can benefit from prevalence differences among routes.
Because the expert family, route strata, calibration layer, and operating-point objective were selected after aggregate target results had been inspected, this extension is neither confirmatory nor protocol-matched to RD25/Reptile.

\clearpage\newpage

\section{Glossary}
\label[appendix]{secGlossary}
Following is a list of notations and acronyms used.
\setglossarystyle{alttree}\glssetwidest{AAAAAAA}
\glsaddallunused[symbols]
\printnoidxglossary[type={symbols}]
\printnoidxglossary[type={acronym}]
\end{document}